\documentclass[lettersize,journal]{IEEEtran}
\usepackage{amsmath,amsfonts}
\usepackage{algorithmic}
\usepackage{algorithm}
\usepackage{array}
\usepackage[caption=false,font=normalsize,labelfont=sf,textfont=sf]{subfig}
\usepackage{textcomp}
\usepackage{stfloats}
\usepackage{url}
\usepackage{verbatim}
\usepackage{graphicx}
\usepackage{multirow}
\usepackage{cite}
\usepackage{bbm}
\newtheorem{cor}{Corollary}
\newtheorem{prop}{Proposition}
\newtheorem{proof}{Proof}
\begin{document}

\title{FedDKD: Federated Learning with \\ Decentralized Knowledge Distillation}
\author{Xinjia~Li, Boyu~Chen and Wenlian~Lu$^*$ \\
\texttt{\{20110180037, 17110180037, wenlian\}@fudan.edu.cn } 
\thanks{$^*$ \textit{Corresponding author.}}
}

\maketitle

\begin{abstract}
The performance of federated learning in neural networks is generally influenced by the heterogeneity of the data distribution. For a well-performing global model, taking a weighted average of the local models, as done by most existing federated learning algorithms, may not guarantee consistency with local models in the space of neural network maps. In this paper, we propose a novel framework of federated learning equipped with the process of decentralized knowledge distillation (FedDKD) (i.e., without data on the server). The FedDKD introduces a module of decentralized knowledge distillation (DKD) to
distill the knowledge of the local models to train the global model by approaching the neural network map average based on the metric of divergence defined in the loss function, other than only averaging parameters as done in literature.
Numeric experiments on various heterogeneous datasets reveal that FedDKD outperforms the state-of-the-art methods with more efficient communication and training in a few DKD steps, especially on some extremely heterogeneous datasets.
\end{abstract}


\section{Introduction}
Data privacy and data security have attracted increasing attention with the widespread application of deep learning in real life, such as smartphones, Internet of Things (IoT)  devices, and digital health. Thus, federated learning \cite{FedAvg, FLLi,2015federated}, which focuses on the challenge of training a global model while leaving the users' private data on their device, has been widely used to train models across mobile phones and IoT devices.

The typical methods of federated learning including FedAvg \cite{FedAvg}, FedProx \cite{FedProx}, and FedMAX \cite{fedMAX}, follow the pipeline proposed by FedAvg. The server collects parameters or gradients from local models and updates the weights of the global model iteratively with the weighted average of the local models. Generally, federated learning employs stochastic gradient descent (SGD) to optimize the local models on the clients. It is well known that the SGD requests independently identically distributed samples to guarantee the stochastic gradient to be an unbiased estimate of the full gradient \cite{FLNonIID,LargeScaleSGD,GD12}.
However, data subsets on each client are heterogeneously distributed in many situations,  where the client datasets have different data sources (e.g., medical radiology images  from various hospitals \cite{FedBN}), or datasets may come from the same source but have a heterogeneous label distribution \cite{FedAvg,FedProx,FedMA,FLNonIID,EnsembleDistillation}.

Previous literature has shown that the performance of federated learning is significantly reduced by data heterogeneity \cite{FLNonIID}, and diverse methods have been proposed to reduce the difference in local models in the parameter space and strengthen the local training on each client. FedProx proposed a proximal term for each local loss to weaken the influence of the local bias, and FedMAX used the Kullback-Leibler (KL) divergence between the activation vector and uniform distribution over the activation vectors, where the activation vector in their work is the last fully connected layer. However, some experiments \cite{FedMA,AsynchronousOF} have revealed that the FedMAX and FedProx perform worse than the FedAvg on several federated learning tasks, reflecting that adding a penalty term in local training is  inefficient enough to solve the heterogeneity problem.

Beyond that, knowledge distillation (KD) has recently been used to reduce the difference in outputs between local and global models. Some researchers \cite{2020Mix2FLD,2020FedED} have assumed that a public dataset is available on the central server to optimize the global model. In particular, FedDF   \cite{EnsembleDistillation} stores unlabeled datasets or artificially generates datasets on the server and
teaches the global model with the ensemble model from clients based on these data. However, the public dataset on the server is not always available or consistent with clients in applications.

Other research, such as FedDistill \cite{fedistill} and FedGen \cite{2021FedGen}, has introduced inductive bias on local models by considering the loss involved other client information and followed FedAvg to obtain global parameters. These data-free methods do not need any public dataset and attempt to mitigate the effects of data heterogeneity only in local models, whose effectiveness is an essential prerequisite for the global model. Besides, FedGen requires an extra well-designed generator in the server, of which the effectiveness greatly determines how well the global model performs. Recently, FedNTD \cite{FedNTD} and FedGKD \cite{FedGDK} distill the knowledge of global model for local models, which add the an extra  distillation loss when training in the local clients.

In addition, several studies have attempted to eliminate the heterogeneity of local datasets by re-balancing the local data distribution. In Astraea \cite{2020Astraea}, they proposed a mediator, grouping the clients to re-balance the probability distributions of local clients and achieve partial equilibrium. A more direct method is distributing a small subset of global data containing a uniform distribution over classes from the cloud to the clients \cite{FLNonIID}. However, these methods require detailed knowledge of the local distribution or available central datasets.

Due to the overparameterization  of neural networks, it has been argued that averaging the parameters is not an optimal approach to averaging the local model in the function space. For instance, in FedMA \cite{FedMA}, the authors presented the argument that the permutation invariance of neural network architectures partially explains why averaging the local parameters is a naive and not optimal method to obtain global parameters, despite its practical performance \cite{FedAvg, fedMAX}. Consider a basic fully connected neural network  $\hat{y}=\sigma(xW^{(1)}\Pi)\Pi W^{(2)}$, where $\{W^{(1)}, W^{(2)}\}$ are optimal weights, and $\Pi$ is a permutation matrix \cite{FedMA}. If both local models respectively obtain the optimal weights $\{W^{(1)}\Pi_{j}, \Pi_{j}W^{(2)}\}$ and  $\{W^{(1)}\Pi_{j'}, \Pi_{j'}W^{(2)}\}$, then averaging the weights is almost impossible to obtain optimal weights given that $\Pi_j\neq \Pi_{j'}$.
However, optimal function mapping still results if the two local models are averaged in the function space (i.e., averaging the same two function mappings in this example).

\begin{figure}
    \centering
    \includegraphics[width=\columnwidth]{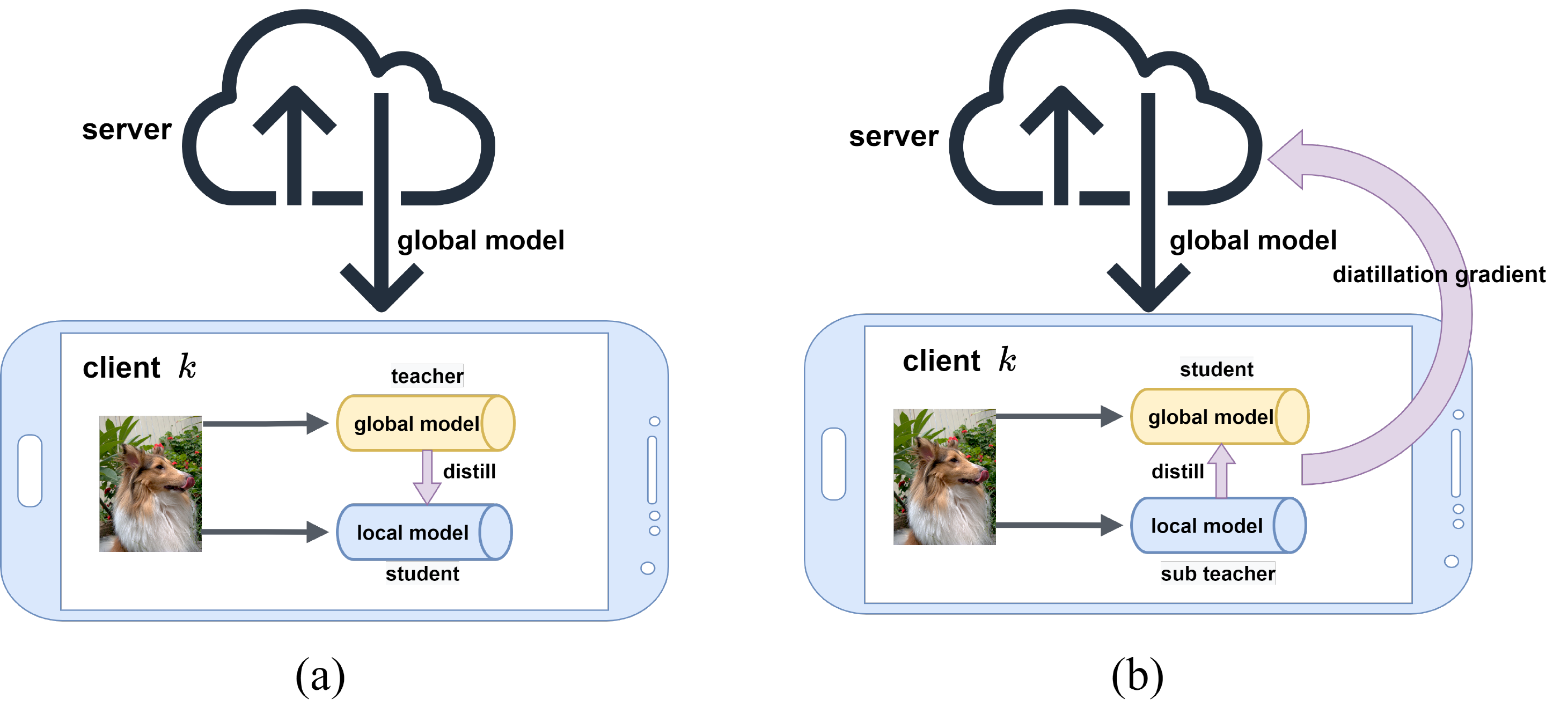}
    \caption{The flowcharts of federated distillation methods without public dataset. Different from the previous methods to distill  the knowledge of the global model for the local models separately (Fig. (a)), our method distills the knowledge of the local models jointly for the global model, which averages the local models in the function space rather than in the parameter space (Fig. (b)).}
    \label{fig::flowchart}
\end{figure}

In this paper, we propose a novel federated learning scheme, \textit{FedDKD}, introducing a module of decentralized knowledge distillation (DKD) to average the local models on the function space instead of the parameter space. Without shared data on the server, the DKD enables the global model to integrate knowledge learned from local datasets to approach the average of local models in the function space. Unlike the existing federated learning methods with knowledge distillation, FedDKD is a data-free method, which means there is no public datasets on the server and does not require an extra generator to generate artifical datasets, optimizing the local models and DKD module alternately. Different from other data-free federated learning methods based on distillation learning which distill the knowledge of the global model for the local models separately (Fig. \ref{fig::flowchart}(a)), FedDKD jointly distills the knowledge of the local models for global model by collecting the knowledge from the clients and updates the global model on the server (Fig. \ref{fig::flowchart} (b)) to obtain the average in function space rather than that in parameter space. The FedDKD can efficiently eliminate the damage caused by the heterogeneity of local datasets and help determine a better global model in the function space compared to directly taking the weighted average of local models.

Numerical experiments on diverse heterogeneous datasets demonstrate the effectiveness of this method to achieve better test accuracy. Moreover, FedDKD achieves training and communication efficiency on some extremely heterogeneous datasets. The DKD module can work as an additional plug-in module for diverse state-of-the-art federated learning technologies to improve their performance.

\section{Decentralized Knowledge Distillation Module}\label{sec::DKD}
Given $K$ clients, where each client $k$ has a private dataset $\mathcal{D}_k$ with distribution $P_k(x)$, where $k=1,\cdots,K$. The empirical distribution function of the whole dataset is denoted as follows:
$$
    P(x) = \sum_{k=1}^K q_k P_k(x),
$$
where $q_k, k=1,2,\cdots K$ is the probability of that $x$ is drawn from the distribution $P_k(x)$ on the client $k$, and $\sum_{k=1}^Kq_k=1$.
The data heterogeneity is caused by the difference among the distributions $\{P_1(x),P_2(x),\cdots,P_K(x)\}$.

Assume that a global ground truth map $\Phi^*(x)$ exists, providing the ground truth label $y$ (i.e.,  $y =\Phi^*(x)$). We attempt to build a parametric global model $\Phi(x; w)$ with parameter $w$ to approximate the ideal $\Phi^*(x)$ by optimizing the following objective function:
\begin{equation}
    \label{equ::flobjective}
      \min_{w} \int div(\Phi^*(x), \Phi(x;w))P(dx),
\end{equation}
where $div(\cdot,\cdot)$ refers to some divergence to measure the distance between $\Phi(x;w)$ and the ideal $\Phi^*(x)$, such as total variation distance $\delta(\cdot, \cdot)$.
\par

In the framework of federated learning, local models $\Phi(x;w_k)$, $k=1,\cdots,K$, on clients are trained by dataset $\mathcal D_{k}$, which have the same structure as the global model $\Phi(x;w)$.
We set the integrated model on clients as follows
\begin{equation}
    \mathcal{T}(x): = \sum_{k=1}^K \Phi(x;w_k) \mathbbm{1}_{\mathcal D_{k}}(x).
\end{equation}
When the triangular inequality holds for some $div(\cdot,\cdot)$ (i.e., $div(u,v)\le div(u,w)+div(w,v)$ for all $u,v$, and $w$ permitted),  we can provide the upper bound of the objective function Eq. (\ref{equ::flobjective}) as follows:
\begin{equation}\label{equ::triangular}
    \begin{aligned}
        & \int div(\Phi^*(x), \Phi(x;w))P(dx) \\
        & \leq \int (div(\Phi^*(x), \mathcal{T}(x))+  div(\mathcal{T}(x), \Phi(x;w)))P(dx) \\
        & = \underbrace{\sum_{k=1}^K q_k \int_{\mathcal D_{k}}  div(\Phi^*(x), \Phi(x;w_k))P_k(dx)}_{\text{$L_{1}$}} \\
        & \underbrace{+\sum_{k=1}^K q_k \int_{\mathcal D_{k}}  div(\Phi(x;w_k), \Phi(x;w))P_k(dx)}_{\text{$L_{2}$}}
    \end{aligned}
\end{equation}
\par
If the triangular inequality does not hold for some divergence (e.g, the KL divergence \cite{1951KL} in classification tasks), Proposition 1 theoretically guarantees that a triangular upper bound exists based on the KL divergence with Pinsker’s inequality\cite{books/daglib/0035708}.
\par
\begin{prop}\label{prop::1}
Define divergence $\delta^2(\cdot, \cdot)$ as $\delta^\dag(\cdot, \cdot)$, where $\delta(\cdot, \cdot)$ is the total variation distance. By Pinsker’s inequality, Eq. (\ref{equ::flobjective}) satisfies the following
\begin{equation}
    \int \delta^\dag(\Phi^*(x), \Phi(x;w))P(dx) \leq L'_1 + L'_2,
\label{equ::deltadiv}
\end{equation}
where
\begin{equation}
    \begin{aligned}
    & L'_1 = \sum_{k=1}^K q_k \int_{\mathcal D_{k}} D_{KL}(\Phi^*(x)|| \Phi(x;w_k))P_k(dx),\\
    & L'_2 = \sum_{k=1}^K q_k \int_{\mathcal D_{k}} D_{KL}(\Phi(x;w_k)|| \Phi(x;w))P_k(dx)
    \end{aligned}
\label{equ::l1l2}
\end{equation}
and $D_{KL}(\cdot||\cdot)$ is the KL divergence. The proof is provided in Supplementary Materials \ref{SM::TVD}.
\end{prop}

\textbf{Motivation.} The general inequality above indicates that minimizing the triangular upper bounds is reasonable instead of minimizing Eq. \eqref{equ::flobjective} in federated learning. Specifically,  minimizing $L_{1}$ in  Eq. (\ref{equ::triangular}) is equivalent to the following
\begin{eqnarray*}
\min_{w_k}\int_{\mathcal D_{k}}div(\Phi(x;w_k), \Phi^{*}(x))P_{k}(dx),~k=1,\cdots,K,
\end{eqnarray*}
 that trains the local model on client $k$ with the private dataset $\mathcal D_{k}$ while minimizing $L_{2}$ in  Eq. (\ref{equ::triangular}) is  the DKD process where the estimated global model $\Phi(x;w)$ learns the knowledge of local trained models $\Phi(x;w_k)$, $k=1,\cdots,K$:
\begin{equation}
\hat{w} = \arg\min_{w}  L_2
\label{equ::dkd}
\end{equation}
that provides the center of all local models $\Phi(\cdot;w_k)$ in the function space.

\textbf{Relation to the weighted averaging method.} Taking the weighted average of the local models (i.e., $\hat{w} \simeq \sum_k^K q_k w_k$)  as the approximation of $\hat{w}$ in Eq. (\ref{equ::dkd}) is a mainstream method in many federate learning algorithms. We analyze its rationality in the homogeneous dataset in terms of the DKD module and demonstrate why it does not work for a heterogeneous dataset.

Define a second-order Taylor approximation of the optimized function in Eq.(\ref{equ::dkd}) on a fixed point $w^*$ as follows:
\begin{equation}
	\begin{aligned}
		& \sum_{k=1}^{K} q_k \int_{\mathcal{D}_k}div(\Phi(x; w_k), \Phi(x; w))P_k(dx) \\
		= & \frac{1}{2}\sum_{k=1}^{K} q_k \langle w_k - w\vert H_k(w^*) \vert w_k - w \rangle \\
		&+ O(\|w_k - w^*\|^3) + O(\|w - w^*\|^3),
	\end{aligned}\label{approx}
\end{equation}
where  $H_k(w^*)=\int_{\mathcal{D}_k}\frac{\partial \Phi(x; w^*)}{\partial w^*}^{\top} H_{\Phi}(x; w^*)\frac{\partial \Phi(x; w^*)}{\partial w^*} P_k(dx)$ and $H_{\Phi}(x; w^*) = \frac{\partial^2 div(u, v)}{\partial u^2}\vert_{u=v=\Phi(x; w^*)}$. The following proposition indicates the effectiveness of the weighted average model when datasets are homogeneous.
\begin{prop}\label{prop::2}
	Assuming that $div(u, v)$ is differentiable for any possible $u$ and $v$, and the datasets $\mathcal{D}_k,~k=1,2\cdots,K$ on the clients are homogeneous (i.e $H_k(w^*) = H(w^*), \forall k=1,\cdots,K$), then $\sum_k{q_k w_k}$ is one of the solutions of the following
	$$
	 \arg\min_{w} \frac{1}{2}\sum_{k=1}^{K} q_k \langle w_k - w\vert H_k(w^*) \vert w_k - w \rangle.
	$$
\end{prop}

Moreover, for the linear model, Corollary \ref{cor::1} is a linear form of Proposition \ref{prop::2}. The detailed proof of Proposition \ref{prop::2} and Corollary \ref{cor::1} are provided in Supplementary Materials  \ref{SM::proof_theorem1}.
\begin{cor}\label{cor::1}
If  $\Phi(x; w) = \Phi(x; (A, b)) = Ax + b$, where $x\in\mathbbm{R}^n$, $A\in \mathbbm{R}^{m, n}$,$b\in \mathbbm{R}^m$, $div(u, v) = \|u - v\|^2_2$, and $\mathcal{D}_k,~k=1,\cdots,K$ are all homogeneous datasets, then one of the solutions of  Eq. (\ref{equ::dkd}) is that
$
\hat{w} = \sum_{k=1}^K{q_k w_k},
$which is the weighted average of the local models.
\end{cor}

\par
For a heterogeneous dataset, which is a more general case in federated learning, there is a high probability that Proposition \ref{prop::2} and Corollary \ref{cor::1} do not hold because $H_k(w^*), k=1,2,\cdots,K$ may be very different. As a result, for heterogeneous datasets, taking the weighted average of local models can not guarantee the optimality of the global model. Thus, we need an algorithm to minimize $L_2$ in Eq.  (\ref{equ::triangular}) to determine a better global model.

\section{The Proposed Approach: FedDKD}
We consider a typical federated learning problem with the disjoint datasets $\mathcal{D}_k:=\{(x_{k,i}, y_{k,i})\}_{i=1}^{n_k}$, where $(x_{k,i}, y_{k,i}) \in \mathbbm{R}^s\times \mathbbm{R}$
on client $k$ . The local model $\Phi(x;w_k)$ on client $k$ and global model $\Phi(x;w)$ are defined with the same neural network architecture.
\subsection{FL with DKD: FedDKD}

To learn the central mapping of local models in function space, we specify the divergence in Eq.  (\ref{equ::flobjective}) as a variant of total variation distance  to measure the distribution distance between the ground truth and prediction of the global model.

Proposition \ref{prop::1} in Section \ref{sec::DKD} notes that with the square of the measurement of total variation distance, Eq. (\ref{equ::flobjective}) can be optimized by minimizing the upper bound $L'_1+L'_2$ based on the KL divergence, from which we can easily optimize a classifier with the derived cross-entropy loss.  Minimizing $L_1'$ trains the local models in parallel on client $k$, whereas minimizing $L_2'$ optimizes the objective in the DKD module to teach the global model using well-trained local models.

Motivated by the above, we propose a federated learning framework equipped with DKD, named \textit{FedDKD}, composed of two stages per round. Each client is trained locally on private data and ground truth labels ($\min L_{1}'$). Then the global model learns from local models in the DKD module, for which the corresponding objective is formulated as follows:
 \begin{equation}
    \label{equ::disObj0}
   \sum_{k\in \mathcal{C}} \frac{q_k}{n_k}\sum_{(x,y)\in \mathcal{D}_k} L_{CE}(\Phi(x;w_k), \Phi(x;w)),
 \end{equation}
 where $\sum_{k\in \mathcal{C}}q_k=1$,  and $L_{CE}(\cdot, \cdot)$ is the cross-entropy loss function.
We define the process to minimize $L_{1}'$ followed by minimizing $L_2'$ as a \textit{DKD round}. When it does not cause ambiguity, we call it a \textit{round} for short.

The optimal solution of minimizing Eq. (\ref{equ::disObj0}) provides the center of all client models $\Phi(x;w_k)$, $k=1,\cdots,K$, in the function space, rather than the center of the parameters of the local models in the parameter space according to the (possibly weighted) Euclidean space.

\begin{algorithm}[tb] 
\caption{\textit{FedDKD}}
\label{algorithm::FedDKD}
\textbf{Input}:Private datasets $\mathcal{D}_k, k=1,2,\cdots, K$ \\
\textbf{Federated learning parameters}:the local epochs $E$, local mini-batch size $B$, the fraction of clients involved  each round $C$, local learning rate $\eta$, the DKD rounds $T$ \\
\textbf{DKD parameters}:
 DKD steps $J$,  DKD learning rate $\gamma$, DKD mini-batch size $\Tilde{B}$ \\
\textbf{Output}: The updated global model $w^*_T$ \\
\textbf{procedure} SERVER ; \par
\begin{algorithmic}[1]
\FOR{Each DKD round $t=1,\cdots, T$}
\STATE
clients $m \longleftarrow  \max(C\cdot K, 1)$
\STATE
Randomly sample a set of $m$ clients $\mathcal{C}_t$
\STATE
For each client $k\in \mathcal{C}_t$, in parallel do
\STATE
$w_k$ $\longleftarrow$ Client-LocalTrain$(k, w^{*}_{t-1})$
\STATE
Get the initial global parameter \\ $\bar{w}_{t,0}=\sum_{k\in \mathcal{C}_t} \frac{n_k}{\sum_{i\in \mathcal{C}_t} n_i} w_k$
\FOR{DKD step $j\in \{1,\cdots, J\}$}
\STATE
For each client $k\in \mathcal{C}_t$, in parallel do
\STATE
        $\nabla d_{k,j}$ $\longleftarrow$ \textit{DKDSGD}$(\bar{w}_{t,j-1}, w_k, \mathcal{D}_k)$
\STATE
        $\bar{w}_{t,j} = \bar{w}_{t,j-1} - \gamma (q_k\sum_{k \in \mathcal{C}_t} \nabla d_{k,j})$
\ENDFOR
\STATE
$w^{*}_t=\bar{w}_{t, J}$
\ENDFOR
\STATE
\textbf{return} $w_T^*$
\end{algorithmic}
\textbf{Client-LocalTrain}$(k,w)$:
\begin{algorithmic}[1]
\STATE
$\mathcal{B} \longleftarrow$ split $\mathcal{D}_k$ into batches of size $B$
\FOR{each local epoch $i=1,2,\cdots E$}
\FOR {batch $b$ in $\mathcal{D}_k$}
\STATE
$w \longleftarrow \text{optimizer}(b;w;\eta)$
\ENDFOR
\ENDFOR
\STATE
\textbf{return} $w$ to server
\end{algorithmic}
\end{algorithm}

 \textbf{Pipeline of FedDKD}. We summarize the
 training pipeline of FedDKD in Algorithm \ref{algorithm::FedDKD}.
In each DKD round $t$, we first sample a subset $\mathcal{C}_t$ of clients, of which the parameters are initialized with $w_{t-1}^*$ (i.e., a global model estimation in the last DKD round). Then, we optimize the local models using local data in their clients in parallel for $E$ local epochs. Afterward, with fixed local models, taking their weighted average, the global model is updated by minimizing Eq. (\ref{equ::disObj0}) for $J$ DKD steps to approach the ideal global model.

 In detail, with the empirical distribution of the sampled local dataset $\mathcal{D}_k$ in client $k$, we use the SGD to estimate the gradient of the sub-term $\frac{1}{n_k}\sum_{(x,y)\in \mathcal{D}_k}L_{CE}(\Phi(x;w_k), \Phi(x;w))$. The algorithm to calculate the gradient on each client is \textit{DKDSGD} (Algorithm \ref{algorithm::DKDSGD}). In each DKD step $j$,  the local model acts as a sub-teacher. In addition, DKDSGD calculates the gradients of $\bar{w}_{t,j}$ on the mini-batch drawn from the local distribution, and sends the gradients to the server. Let $q_k=1/|\mathcal{C}_t|, k\in \mathcal{C}_t$, and the server collects all gradients $\nabla d_{k,j}, k\in \mathcal{C}_t$ from local clients  and averages them to obtain the  gradient of Eq. (\ref{equ::disObj0}) concerning $\bar{w}_{t,j}$. After acquiring the  gradient for Eq. (\ref{equ::disObj0}), the global model updates with the DKD learning rate $\gamma$ and distributes the new global parameters to the local clients. In FedDKD  (Algorithm \ref{algorithm::FedDKD}), Lines 7 to 11 represent the DKD module.

\begin{algorithm}[tb] 
\caption{\textit{DKDSGD}. Solver with SGD on each client $k$ in step $j$}
\label{algorithm::DKDSGD}
\textbf{Input}:The global parameters $w$, the parameters of client $k$'s model $w_k$, local training dataset $\mathcal{D}_k$; \\
\textbf{Parameter}: DKD mini-batch size $\Tilde{B}$; \\
\textbf{Output}:The local gradient $\nabla d_{k,j}$.

\begin{algorithmic}[1]
\STATE
Sample a min-batch $\mathcal{B}_k$\ from training data $\mathcal{D}_k$ with batch size $\Tilde{B}$ ;
\STATE
Get the logit $\Phi(\mathcal{B}_k; w_k)$ of local model on client $k$ with local parameters $w_k$ and the logit $\Phi(\mathcal{B}_k; w)$ of global model with global parameters $w$.
\STATE
$\nabla d_{k,j} = \frac{1}{\Tilde{B}} \frac{\partial
\sum_{(x,y)\in \mathcal{B}_k}L_{CE}(\Phi(x; w_k), \Phi(x; w))
}{\partial w}$
\STATE
\textbf{return} $\nabla d_{k,j}$
\end{algorithmic}
\end{algorithm}

\textbf{Benefits of FedDKD}.
Minimizing $L_1'$ and $L_2'$ in Eq. (\ref{equ::l1l2}) alternately, FedDKD minimizes the upper bound of the distance between the ground truth map and global model in the function space. The keypoint of FedDKD is minimizing both terms, whereas most of the federated learning algorithms only optimize $L_1'$.
Eliminating the damage caused by the heterogeneity of the local datasets and  reparameterization invariance of neural networks on the map functions by minimizing $L_2'$, the DKD module helps FedDKD search for a better global model for the current DKD round. It sends back better initial parameters for the next DKD round for local training.

\subsection{Extensions of FedDKD}
 We note that the framework of FedDKD can be a general framework of federated learning because the DKD process can be a plug-in for other federated learning methods.
 \par
\textbf{FedDKD\_MAX}. We can replace the local training scheme of FedDKD  with that of FedMAX \cite{fedMAX} to enable similar activation vectors across various clients. In detail, we add a loss term $\beta \frac{1}{B}\sum_{i=1}^N KL(a_i\|U)$ to the cross-entropy loss in local training on the clients, where $a_i$ is the activation vector of sample $i$, $\beta$ is a hyper-parameter, and $U$ is the uniform distribution over the activation vectors. We call the alternative algorithm FedDKD\_MAX.
\par
\textbf{FedDKD\_BN}. We combine FedDKD with FedBN \cite{FedBN} to handle the neural network models with batch normalization (BN) layers, called FedDKD\_BN, where local models have their own BN layers whose parameters should remain. Thus, for FedDKD\_BN, the objective of the DKD module is revised as follows:
$$
	\label{equ::disObjBN}
	\min_{w'\in \mathbbm{R}^d} \sum_{k\in \mathcal{C}} \frac{q_k}{n_k} \sum_{(x,y)\in \mathcal{D}_k} L_{CE}(\Phi_k(x;w_k) , \Phi(x;w',w_k^{BN})),
$$
where $w_k^{BN}$ is the parameter of the BN layers of the local model $k$, and $w'$ represents other parameter. In the DKD module of FedDKD\_BN, only the weight $w'$ is updated. Based on FedBN, FedDKD\_BN takes weighted average parameters, excluding the $BN$ layers, as the initial global model of the DKD module.
Please refer to Supplementary Materials \ref{SM::FedDKD_fedMAX} and \ref{SM::FedDKD_FedBN} for more details on these extensions.
\section{Experiment}
We perform many numerical  experiments on the
EMNIST/FEMNIST \cite{FEMNIST,fedMAX}, CIFAR10/CIFAR100 \cite{cifar-10},  and a multi-sources digits dataset, compared with several mainstream methods, including FedAvg \cite{FedAvg}, FedProx \cite{FedProx}, FedMAX \cite{fedMAX}, and FedBN \cite{FedBN}. An experiment on EMNIST compares several state-of-the-art data-free KD approaches, such as FedGen \cite{2021FedGen} and FedDistill \cite{fedistill}. Overall, we experiment on the following sufficient settings:
\begin{itemize}
    \item various datasets, such as EMNIST, FEMNIST, CIFAR10, CIFAR100, and multi-sources digits datasets;
    \item various heterogeneities: part of the categories, Dirichlet distribution with different $\alpha$ values, and different data sources;
    \item different activation ratios of clients: 0.25, 0.5, 0.75, and 1.0; and
    \item different local epochs.
\end{itemize}
We also experimented with the following sufficient metrics:
\begin{itemize}
    \item test accuracy,
    \item costs to reach the target accuracy,
    \item test accuracy under the same communication cost, and
    \item test accuracy under the same training steps ($\times 10^3$).
    \end{itemize}
In addition, we use different models with or without BN layers.

\subsection{FEMNIST}
In this section, we demonstrate the superiority of the FedDKD in communication efficiency and convergence speed when handling heterogeneity datasets. Moreover, we display the power of the DKD module equipped with other state-of-the-art training techniques for the local models.

We compare FedMAX  with the proposed method in experiments on the dataset FEMNIST \cite{fedMAX}, which is divided into heterogeneous local datasets. We randomly choose $6$ out of $26$ classes to allocate to each agent. We use $50\%$ of the raw training dataset as the federated learning training dataset and  $10\%$ as the validation dataset. The test dataset remains the same as the raw dataset.

We set the number of clients to $20$, and the activation ratio of clients per DKD round is $0.25, 0.5,$ or $0.75$. The number of local training epochs is $5$ or $15$, and the total DKD rounds $T$ is $660$. For FedProx and FedMAX, we perform the optimal search for the best parameters $\mu$ and $\beta$, respectively.
For FedDKD and FedDKD\_MAX, we set DKD step $J$ to 10, and the initial DKD learning rate $\gamma$ is $0.2$ with a decay rate of 0.98 per DKD round.
We run each experiment three times with different random seeds to obtain credible results. The test accuracy is based on the best model for the validation dataset. Please refer to  Supplementary Materials \ref{SM::FEMNIST} for more details.
\par
\textbf{Reaching a target quickly and efficiently}. Three metrics are considered to reach a target test accuracy, including the DKD round, communication round, and average training step per activated client. In addition, FedDKD requires much fewer DKD rounds and training steps in all experiment settings and is efficient in communication when the activation ratio $c$ is 0.5 or 0.75, as the DKD module can estimate the global distribution of the dataset better. For example, if the activation ratio $c=0.75$ and the local training epoch $E=5$, FedDKD can achieve a $90.5\%$ test accuracy with only $5\%$ of the DKD rounds and training steps and $50\%$ of the communication rounds compared to those in other mainstream methods, as Table \ref{tab::FemnistT} presents.  Combining the DKD module with FedMAX, FedDKD\_MAX requires fewer training steps and DKD rounds when $c=0.25$ and $E=5$.
\begingroup
\begin{table*}[!h]
\footnotesize
\caption {\label{tab:table1} Average number of training steps to reach target performance $T$ ($90.5\%$ for $E=5$ and $90\%$ for $E=15$)}
\begin{tabular}{l|l|lll|lll|lll}
\hline
\hline
 \multirow{2}{*}{} & \multirow{2}{*}{$E$} & \multicolumn{3}{c}{$c=0.25$} & \multicolumn{3}{c}{$c=0.5$} & \multicolumn{3}{c}{$c=0.75$}   \\
\cline{3-5} \cline{6-8} \cline{9-11}
 & &  round$^\dag$   & step$^\ddag$  & comm $^\S$     &  round$^\dag$   & step$^\ddag$  & comm $^\S$    &  round$^\dag$   & step$^\ddag$  & comm $^\S$   \\
\hline
\multirow{2}{*}{FedAvg} & 5 & $330\pm 76$ & $81\pm 19$ & $330\pm 76$ & $245\pm 90$ & $60\pm 22$ & $245\pm 90$ & $283\pm 145$ & $69\pm 36$ & $283\pm 145$ \\
& 15 & $353\pm 86$ & $260\pm 63$ & $\mathbf{353\pm 86}$ & $350\pm 98$ & $257\pm 72$ & $350\pm 98$ & $332\pm 115$ & $244\pm 85$ & $332\pm 115$ \\
\hline
\multirow{2}{*}{FedProx} & 5 & $330\pm 76$ & $81\pm 19$ & $330\pm 76$ & $263\pm 112$ & $64\pm 28$ & $263\pm 112$ & $301\pm 145$ & $74\pm 36$ & $301\pm 145$ \\
& 15 & $425\pm 25$ & $313\pm 18$ & $425\pm 25$ & $339\pm 113$ & $249\pm 83$ & $339\pm 113$ & $339\pm 158$ & $249\pm 116$ & $339\pm 158$ \\
\hline
\multirow{2}{*}{FedMAX} & 5 & $310\pm 101$ & $76\pm 25$ & $\mathbf{310\pm 101}$ & $258\pm 73$ & $63\pm 18$ & $258\pm 73$ & $276\pm 105$ & $68\pm 26$ & $276\pm 105$ \\
& 15 & $388\pm 90$ & $285\pm 66$ & $388\pm 90$ & $367\pm 98$ & $270\pm 72$ & $367\pm 98$ & $376\pm 127$ & $276\pm 93$ & $376\pm 127$ \\
\hline
\multirow{2}{*}{FedDKD} & 5 & $98\pm 4$ & $25\pm 1$ & $1074\pm 40$ & $\mathbf{18\pm 3}$ & $\mathbf{5\pm 1}$ & $\mathbf{198\pm 32}$ & $\mathbf{13\pm 2}$ & $\mathbf{3\pm 0}$ & $\mathbf{139\pm 21}$ \\
& 15 & $\mathbf{104\pm 12}$ & $\mathbf{77\pm 9}$ & $1140\pm 129$ & $\mathbf{22\pm 8}$ & $\mathbf{17\pm 6}$ & $\mathbf{246\pm 93}$ & $\mathbf{12\pm 2}$ & $\mathbf{9\pm 2}$ & $\mathbf{132\pm 24}$ \\
\hline
\multirow{2}{*}{FedDKD\_MAX} & 5 & $\mathbf{96\pm 6}$ & $\mathbf{24\pm 2}$ & $1052\pm 70$ & $19\pm 3$ & $5\pm 1$ & $205\pm 36$ & $15\pm 3$ & $4\pm 1$ & $169\pm 29$ \\
& 15 & $151\pm 47$ & $113\pm 35$ & $1665\pm 519$ & $23\pm 8$ & $17\pm 6$ & $257\pm 87$ & $13\pm 2$ & $9\pm 1$ & $139\pm 21$ \\
 \hline
 \hline
\end{tabular}
\begin{tabbing}
$^\dag$ DKD round, \quad $^\ddag$ local training step, \quad $^\S$ communication round. If not mentioned, the followings are the same as that.
\end{tabbing}
\label{tab::FemnistT}
\end{table*}
\endgroup

\textbf{Best performance}. Training for $660$ DKD rounds,  FedDKD and FedDKD\_MAX outperform all the other methods by around $0.75\%, 1.0\%$, and $1.5\%$ on test accuracy for $c=0.25, 0.5$ and $0.75$, respectively (Table \ref{tab::femnistBest}). With more clients activated in each communication round, FedDKD can better estimate the global distribution of the datasets and have more advantages than other methods on test accuracy. To compete with other methods fairly in communication cost, we also present the test accuracy when FedDKD and FedDKD\_MAX are only trained for $60$ DKD rounds. In this setting, all methods communicate for 660 rounds with their local clients, and the communication cost in each round remains the same. However, the average training steps of FedDKD and FedDKD\_MAX is only about $10\%$ of that of the baselines. The results reveal that FedDKD can achieve a better test accuracy with the same training steps and fewer DKD rounds. When the activation ratio $c$ increases, with the same communication cost, FedDKD can achieve better performance with fewer training steps and DKD rounds.
\par
\begin{figure}
    \centering
    \includegraphics[width=0.48\columnwidth]{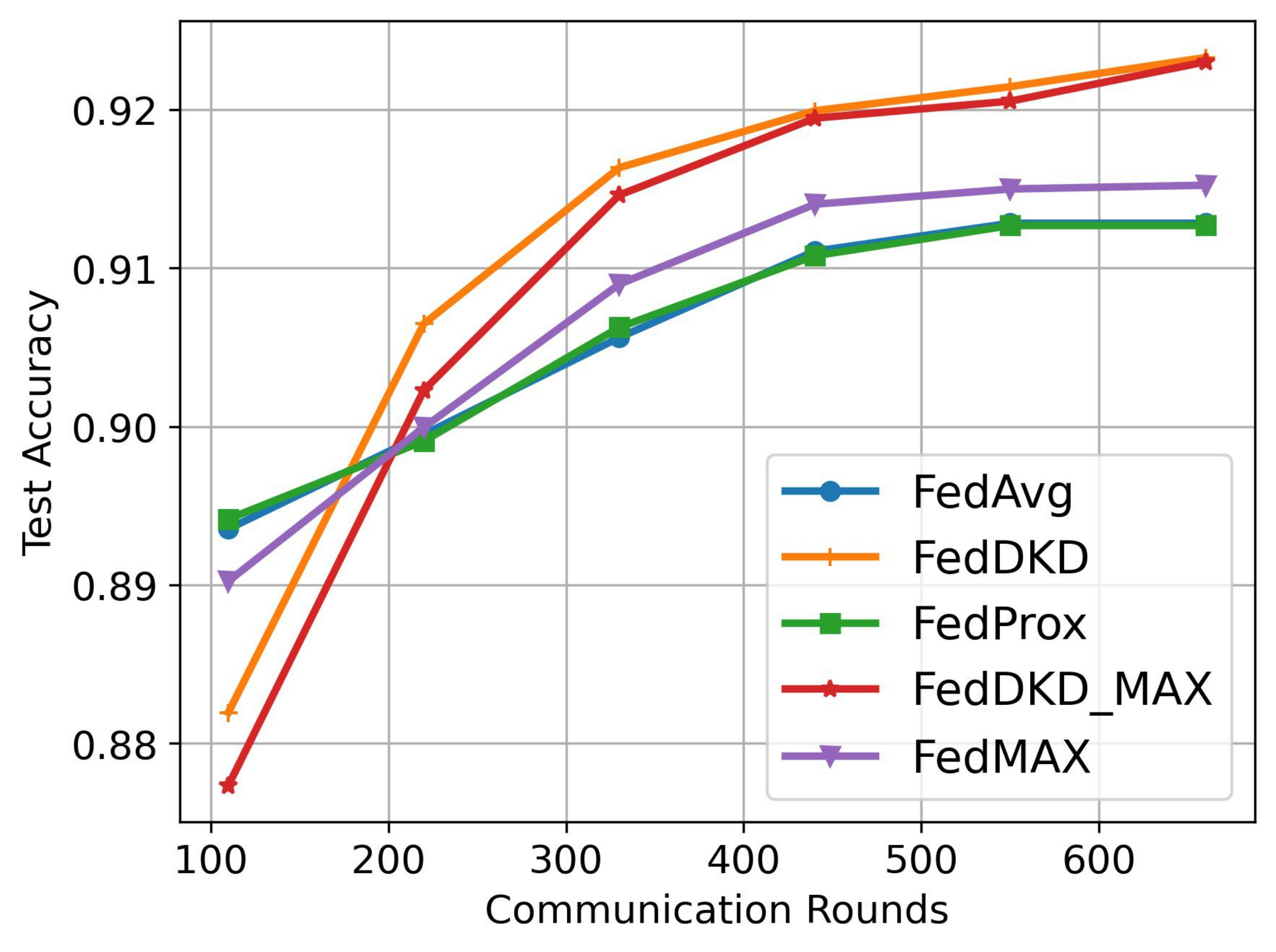}
    \includegraphics[width=0.48\columnwidth]{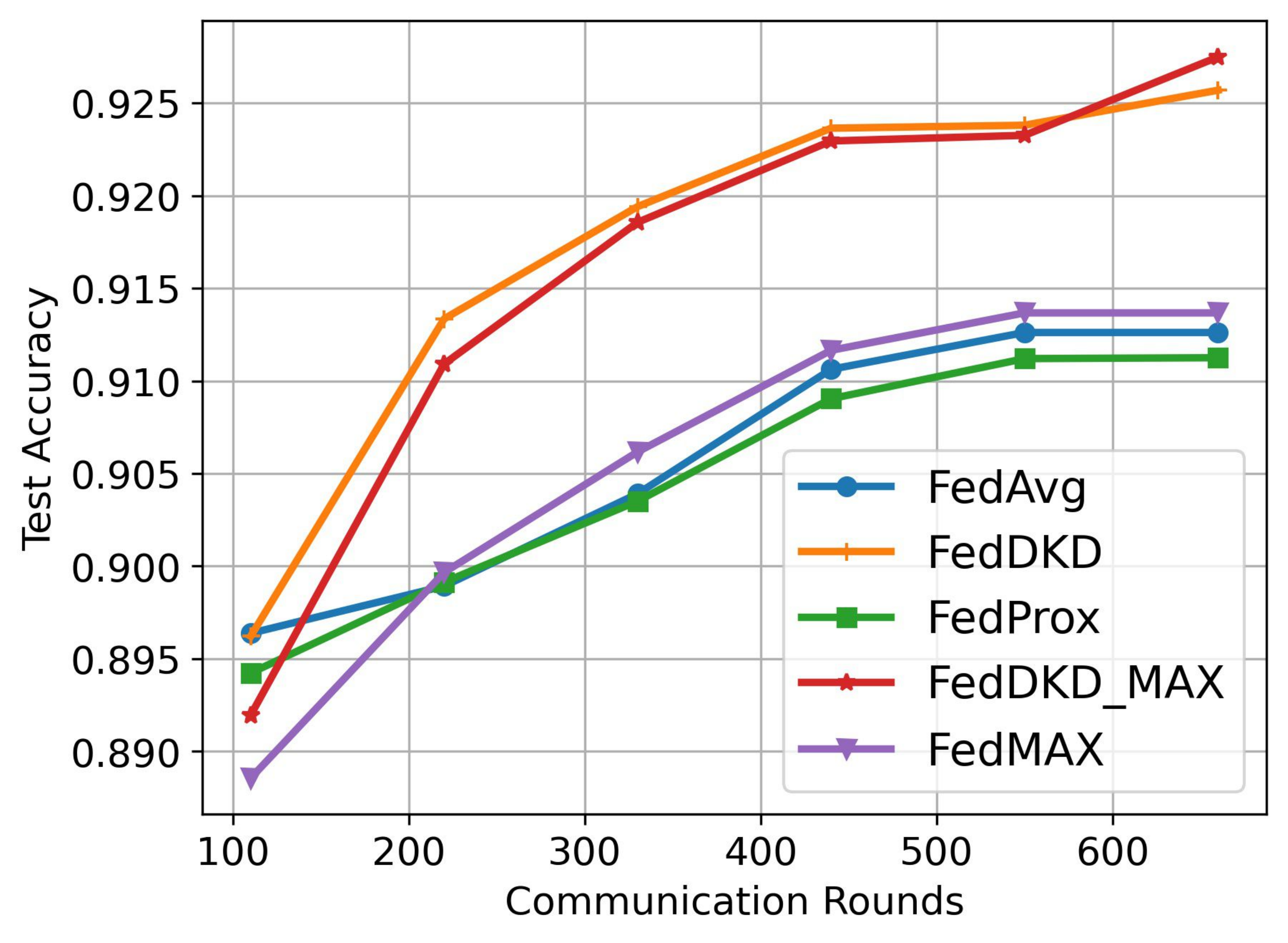}
    \caption{Test accuracy under the same communication cost for $c=0.5$ (left) and $c=0.75$ (right), where $E=5$.}
    \label{fig::comm}
\end{figure}

\textbf{Efficient communication}. The FedDKD method is efficient in communication when the activation ratio $c$ is $0.5$ or $0.75$. With the same communication rounds and communication costs (Figure \ref{fig::comm}), the proposed method achieves the best test accuracy. For detailed results for $E=15$, please refer to Supplementary Materials \ref{SM::FEMNIST}. With the same communication rounds, FedDKD and FedDKD\_MAX only run for $1/11$ of the DKD round and about $10\%$ of the training steps of the baselines. Thus, the proposed methods may be weaker than other methods when the communication rounds are too few.
\par
\begin{figure}
    \centering
    \includegraphics[width=0.48\columnwidth]{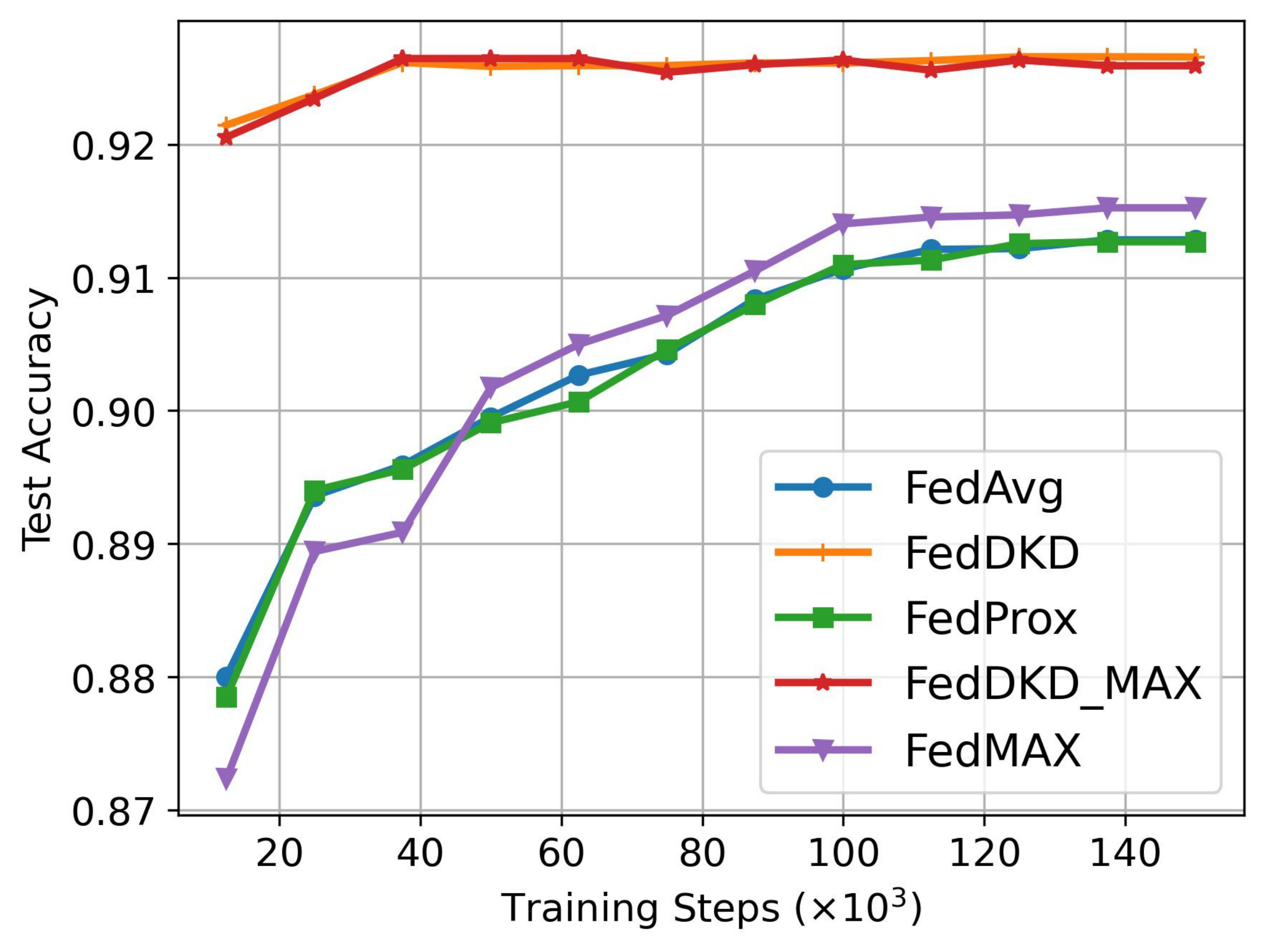}
    \includegraphics[width=0.48\columnwidth]{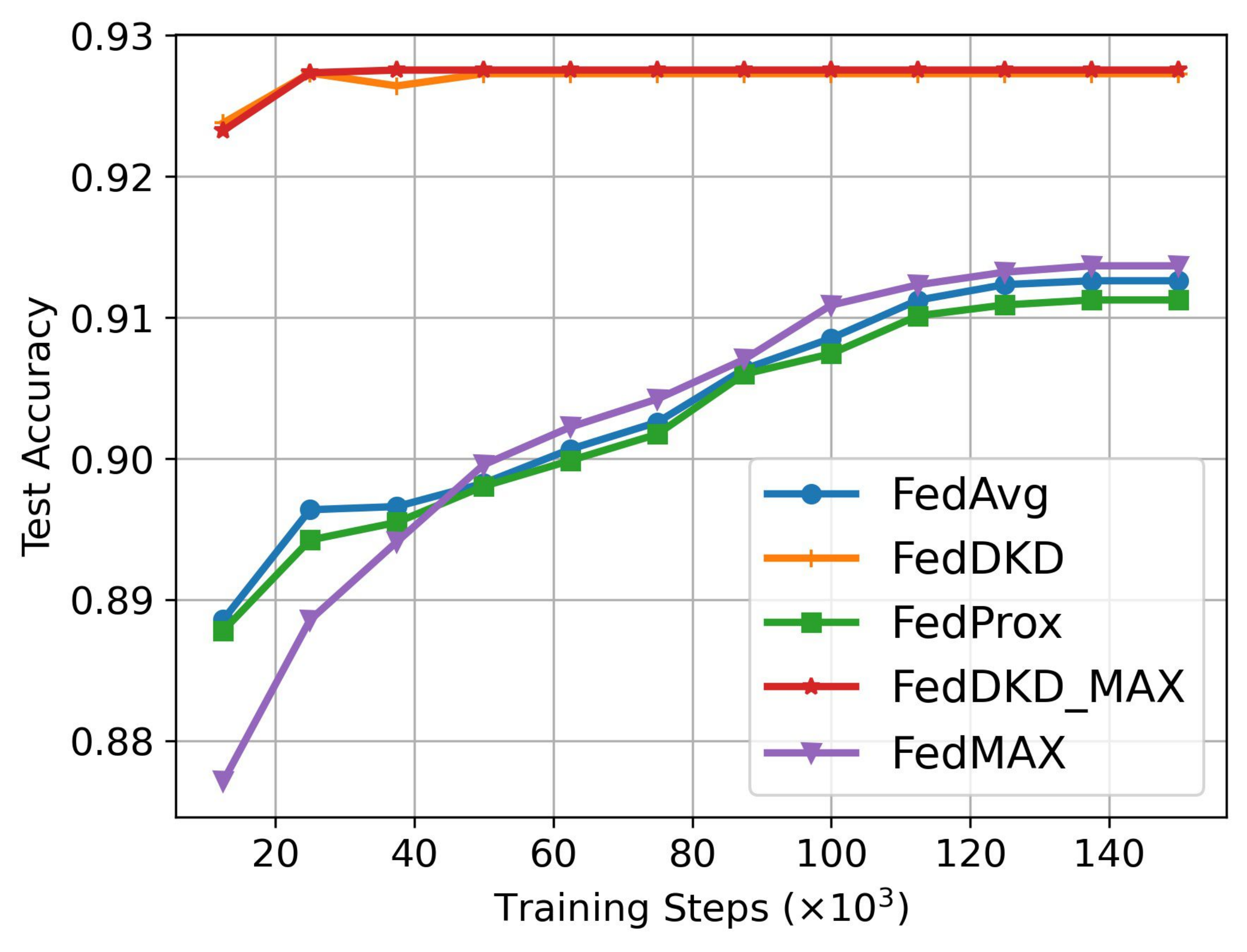}
    \caption{Test accuracy under the same training steps for $c=0.5$ (left) and $c=0.75$ (right), where $E=5$.}
    \label{fig::train_step}
\end{figure}
\textbf{Efficient training}. The proposed methods require a fewer more training step per DKD round in the DKD module. However, the improvement is not due to more training steps, and FedDKD is training-efficient.
With the same training steps, FedDKD and FedDKD\_MAX display a considerable improvement in test accuracy for the baselines (Figure \ref{fig::train_step}).
In comparing the proposed methods and the baselines, many unnecessary training steps occur for the baselines, and more communication is the key factor in pursuing a better performance.
Please refer to Supplementary Materials \ref{SM::FEMNIST} for the results of $E=15$ and $E=5,c=0.25$.

\begingroup
\begin{table*}[!h]
\small
\centering
\caption {\label{tab:table2} Test accuracy on the FEMNIST dataset}
\begin{tabular}{l|l|ll|ll|ll}
\hline
\hline
 \multirow{2}{*}{} & \multirow{2}{*}{$E$} & \multicolumn{2}{c}{$c=0.25$} & \multicolumn{2}{c}{$c=0.5$} & \multicolumn{2}{c}{$c=0.75$}   \\
\cline{3-4} \cline{5-6} \cline{7-8}
 &  & test accuracy   & round  & test accuracy   & round   & test accuracy   & round   \\
\hline
\multirow{2}{*}{FedAvg} & 5 & $0.9156\pm 0.0026$ & $521\pm 20$ &  $0.9128\pm 0.0027$ & $523\pm 16$ &  $0.9126\pm 0.0028$ & $519\pm 16$ \\
& 15 & $0.9114\pm 0.0038$ & $586\pm 43$ &  $0.9069\pm 0.0041$ & $607\pm 16$ & $0.9077\pm 0.0034$ & $589\pm 9$ \\
\hline
\multirow{2}{*}{FedProx} & 5 & $0.9134\pm 0.0045$ & $459\pm 52$ &  $0.9127\pm 0.0043$ & $513\pm 17$ &  $0.9113\pm 0.0029$ & $506\pm 37$ \\
& 15 & $0.9062\pm 0.0036$ & $527\pm 1$ &  $0.9088\pm 0.0032$ & $639\pm 16$ & $0.9067\pm 0.0053$ & $600\pm 43$ \\
\hline
\multirow{2}{*}{FedMAX} & 5 & $0.9158\pm 0.0050$ & $526\pm 14$ &  $0.9152\pm 0.0042$ & $518\pm 29$ &  $0.9137\pm 0.0036$ & $516\pm 7$ \\
& 15 & $0.9098\pm 0.0049$ & $562\pm 29$ &  $0.9073\pm 0.0045$ & $534\pm 28$ & $0.9059\pm 0.0042$ & $570\pm 61$ \\
\hline
\multirow{2}{*}{FedDKD} & 5 & $0.9245\pm 0.0015$ & $460\pm 57$ &  $0.9246\pm 0.0023$ & $271\pm 220$ &  $0.9273\pm 0.0014$ & $113\pm 33$ \\
& 15 & $\mathbf{0.9189\pm 0.0001}$ & $613\pm 56$ &  $\mathbf{0.9200\pm 0.0012}$ & $261\pm 220$ & $\mathbf{0.9227\pm 0.0020}$ & $76\pm 25$ \\
\hline
\multirow{2}{*}{FedDKD\_MAX} & 5 & $\mathbf{0.9251\pm 0.0012}$ & $480\pm 40$ &  $\mathbf{0.9259\pm 0.0011}$ & $238\pm 184$ &  $\mathbf{0.9275\pm 0.0017}$ & $93\pm 27$ \\
& 15 & $0.9174\pm 0.0018$ & $560\pm 31$ &  $0.9194\pm 0.0005$ & $109\pm 17$ & $0.9211\pm 0.0018$ & $67\pm 20$ \\
\hline
\multirow{2}{*}{FedDKD@60} & 5 & $0.8904\pm 0.0078$ & $53\pm 5$ &  $0.9202\pm 0.0014$ & $51\pm 10$ &  $0.9257\pm 0.0006$ & $53\pm 5$ \\
& 15 & $0.8875\pm 0.0116$ & $55\pm 4$ &  $\mathbf{0.9183\pm 0.0012}$ & $52\pm 9$ & $\mathbf{0.9222\pm 0.0026}$ & $40\pm 3$ \\
\hline
\multirow{2}{*}{FedDKD\_MAX@60}$^\dag$ & 5 & $0.8915\pm 0.0140$ & $50\pm 9$ &  $\mathbf{0.9230\pm 0.0024}$ & $51\pm 10$ &  $\mathbf{0.9275\pm 0.0010}$ & $58\pm 1$ \\
& 15 & $0.8795\pm 0.0089$ & $52\pm 5$ &  $0.9165\pm 0.0024$ & $52\pm 8$ & $0.9213\pm 0.0020$ & $50\pm 7$ \\
 \hline
 \hline
\end{tabular}
\begin{tabbing}
$^\dag$Take $60$ DKD rounds.
\end{tabbing}
\label{tab::femnistBest}
\end{table*}
\endgroup
\subsection{CIFAR-10 and CIFAR-100}\label{sec::CIFA10}
\begin{table*}[!h]
\centering
\caption{Performance overview of CIFAR-10}
\begin{tabular}{c|c|c|c|c|c|c|c}
\hline
\multirow{2}{*}{Methods} & \multicolumn{4}{c|}{top-1 test acc} & \multicolumn{3}{c}{test accuracy $\ge$ 74\%} \\ \cline{2-8}& test accuracy & round &comm & train step & round &comm & train step \\
\hline
FedAvg & $0.7531\pm0.0108$ & $348\pm2$ & $348\pm2$ & $154.3\pm0.8$ &$292\pm38$ & $292\pm38$ & $129.4\pm17.0$ \\
\hline
FedProx & $0.7528\pm0.0107$ & $343\pm3$ & $343\pm3$ & $152.4\pm1.5$ &$274\pm53$ & $274\pm53$ & $121.6\pm23.4$ \\
\hline
FedDKD & $\mathbf{0.8015\pm0.0120}$ & $271\pm6$ & $1084\pm23$ & $121.1\pm2.6$ &$\mathbf{55\pm4}$ & $\mathbf{219\pm18}$ & $\mathbf{24.4\pm2.0}$ \\
\hline
FedDKD@87$^\dag$ & $0.7729\pm0.0104$ & $\mathbf{84\pm2}$ & $\mathbf{337\pm10}$ & $3\mathbf{7.7\pm1.1}$ &$\mathbf{55\pm4}$ & $\mathbf{219\pm18}$ & $\mathbf{24.4\pm2.0}$ \\
\hline
\end{tabular}
\label{tab::cifar10}
\flushleft{$^\dag$ Take 87 DKD rounds.}

\end{table*}

\begin{table*}[!h]
\centering
\caption{Performance overview of CIFAR-100}
\begin{tabular}{c|c|c|c|c|c|c|c}
\hline
\multirow{2}{*}{Methods} & \multicolumn{4}{c|}{top-1 test acc} & \multicolumn{3}{c}{test accuracy $\ge$ 51.4\%} \\ \cline{2-8}& test accuracy & round &comm & train step & round &comm  & train step \\
\hline
FedAvg & $0.5188\pm0.0041$ & $339\pm3$ & $339\pm3$ & $150.4\pm1.3$ &$244\pm22$ & $244\pm22$ & $108.1\pm9.6$ \\
\hline
FedProx & $0.5227\pm0.0067$ & $324\pm5$ & $\mathbf{324}\pm\mathbf{5}$ & $143.9\pm2.4$ &$232\pm59$ & $\mathbf{232}\pm\mathbf{59}$ & $103.1\pm26.3$ \\
\hline
FedDKD & $\mathbf{0.5424}\pm\mathbf{0.0130}$ & $274\pm48$ & $1007\pm193$ & $115.1\pm20.5$ &$\mathbf{81}\pm\mathbf{22}$ & $244\pm66$ & $\mathbf{28.8}\pm\mathbf{14.6}$ \\
\hline
FedDKD@110$^\dag$ & $0.5241\pm0.0109$ & $\mathbf{105}\pm\mathbf{4}$ & $330\pm14$ & $\mathbf{39.4}\pm\mathbf{7.0}$ &$\mathbf{81}\pm\mathbf{22}$ & $244\pm66$ & $\mathbf{28.8}\pm\mathbf{14.6}$ \\
\hline
\end{tabular}
\label{tab::cifar100}
\flushleft{$^\dag$ Take 110 DKD rounds.}
\end{table*}
In this experiment, we consider a more challenging task of the datasets CIFAR-10 and CIFAR-100 \cite{cifar-10}. Following the previous work \cite{FedMA}, we divide CIFAR-10 into $K$ clients by sampling $\textbf{p}_r \sim \text{Dir}_K(\alpha)$ from a Dirichlet distribution and allocating a $\textbf{p}_{r,k}$ proportion of the training instances of class $r$ to the local client $k$. In the experiments, $\alpha=0.1$, and the local datasets are extremely heterogeneous. We use the same neural network model (i.e. VGG-9) without BN layers, as in \cite{FedMA}. We set $10\%$ of the raw training set as the validation set and the others as the training set.

For federated learning experiments, we set the number of clients to $16$, and the client fraction $C$ is $1.0$ (i.e., all clients will be involved in the local update per DKD round). To achieve a trade-off between performance and communication costs, we set DKD step J to  $3$. The DKD learning rate $\eta_{d}$ is $0.08$ with a decay rate of $0.99$ per DKD round for CIFAR-10 and $\eta_{d}=0.06$ for CIFAR-100.  The test accuracy and corresponding DKD  and communication rounds are based on the best model for the validation set. Please refer to Supplementary Materials \ref{SM::CIFAR10} for the details of experiment settings.

For CIFAR-10 (Table \ref{tab::cifar10}), with  $350$ DKD rounds, the proposed FedDKD improves the test performance by almost $5\%$ of the baselines at the expense of similar training costs and  only four times the communication cost. With a similar communication cost and much fewer training steps (i.e., $24.4\%$ of the training steps of the baselines), FedDKD achieves almost $2\%$ improvement on the test accuracy in $87$ DKD rounds. To reach the same test accuracy target (i.e., $74\%$) with less communication cost, FedDKD requires only about $20\%$ of the DKD rounds and $20\%$ of the training steps of the baseline cost.

For CIFAR-100 (Table \ref{tab::cifar100}), where it is more challenging to handle heterogeneity because the number of categories is $100$, local teachers have low test accuracy (less than $40\%$) in the first $30$ DKD rounds, and the knowledge is too incomplete to teach the student model. Thus, the DKD step is not executed for the first $30$ DKD rounds in this experiment. In addition, FedDKD improves the global model by about $2\%$ regarding test accuracy. It only needs $1/3$ of the DKD round and  training steps in other methods to reach the same test accuracy target (i.e., 51.4\%) under the same communications, which illustrates the training efficiency of the proposed method. For a fair comparison, we also run FedDKD for 110 DKD rounds (i.e., FedDKD@110), where the number of communication rounds is $350$. In this case, FedDKD still slightly exceeds the baseline in test accuracy. However, FedDKD only trains about $1/3$ of the steps.

The experiments on CIFAR-10/100 illustrate that the proposed method is efficient in much more complex tasks with equal DKD rounds and training steps. With equal communication costs and much fewer training steps, FedDKD exceeds the baselines on test accuracy. The experiments imply that local training is often superfluous for heterogeneous datasets and that more communication is the key to higher performance.

\subsection{Knowledge Distillation Algorithms}
This subsection compares FedDKD with other data-free federated learning algorithms equipped with KD, such as FedGen and FedDistill.

In FedGen, the server trains a generator to ensemble the client information in a data-free manner and then broadcast to users, regulating local training using the learned knowledge as an
inductive bias. Additionally, FedDistill only collects and sends back the average logits  of the client models. Following previous research \cite{2021FedGen}, we also share the parameters of the client models as FedAvg for FedDistill, called FedDistill$^+$ to achieve a fair comparison.

The experiments on the EMNIST dataset follow the work in \cite{2021FedGen}. There are $20$ clients with an active ratio of $50\%$. The local epoch $E=20$. We use only $10\%$ of the training dataset and all test datasets. To obtain an extremely heterogeneous  dataset, we divide the training dataset according to the Dirichlet distribution with $\alpha=0.05$ or $0.1$. We maintain the hyper-parameters for the baseline as those in FedGen \cite{2021FedGen}. For FedDKD, we set the DKD learning rate $\gamma=0.40$ with a decay rate of $0.99$ per DKD round and the DKD steps $J=3$. We run each experiment $10$ times.

\begin{table}[h]
\centering
\caption{Test accuracy on the EMNIST dataset}
\begin{tabular}{c|c|c}
\hline
Methods  & $\alpha=0.05$ &  $\alpha=0.1$ \\
\hline
FedAvg  & $64.76\pm2.11$  & $69.07\pm 1.71$ \\
\hline
FedProx &  $63.90\pm 2.13$ &  $68.49\pm 1.74$ \\
\hline
FedDistill$^+$ & $63.18\pm 2.31$  & $69.20\pm 1.44$ \\
\hline
FedGEN  & $68.56\pm 1.82$ &  $72.06\pm 1.61$ \\
\hline
FedDKD  & $\textbf{72.77}\pm \textbf{1.38}$ &  $\textbf{73.75}\pm \textbf{1.60}$ \\
\hline
\end{tabular}
\label{tab::EMNIST}
\end{table}
As Table \ref{tab::EMNIST} lists, FedDKD achieves the best test accuracy and outperforms other methods by about $4.2\%$ when $\alpha=0.05$ and by $1.7\%$ when $\alpha=0.1$. In addition,  FedDKD does not require a well-designed and well-trained extra generator as in FedGEN but achieves superb performance when the dataset is highly  heterogeneous.
\begin{table*}[t]
    \centering
    \caption{The results for the multi-sources digits dataset in $100$ DKD rounds.}
    \begin{tabular}{c|c | c| c| c | c}
    \hline
         \textbf{Method} & FedAvg & FedProx & FedBN & FedDKD & FedDKD\_BN \\
         \hline
         best test accuracy & $82.58\% \pm 0.33\%$ & $82.58\%\pm 0.40\%$ & $85.48\% \pm0.20\%$  & $85.24\%\pm 0.17\%  $ & $\mathbf{85.80\% \pm 0.18\%} $ \\
         \hline
         final test accuracy & $82.32\%\pm 0.22\%$ &  $82.37\%\pm 0.20\%$  & $85.40\%\pm 0.26\%$ & $85.01\%\pm 0.23\%$ & $\mathbf{85.71\%\pm 0.17\%}$ \\
         \hline
    \end{tabular}
    \label{tab::FedBN_results}
\end{table*}
\subsection{Architecture with the Batch Normalization Layer }
In this experiment, we consider the architecture with the BN layer. Most models remove the BN layer in federated learning to average the model better. However, FedBN introduces the BN layer to handle heterogeneous datasets when the local datasets are collected from different sources. This subsection primarily demonstrates the power of the DKD module equipped with other state-of-the-art methods (i.e., FedBN).

Specifically, we use five different digit datasets: SVHN \cite{SVHN}, USPS \cite{USPS}, SynthDigits \cite{SynthDigits-MNIST-M}, MNIST-M \cite{SynthDigits-MNIST-M}, and MNIST \cite{MNIST}, with 10\% of each dataset allocated to each client with $743$ training samples for each local model. We use the same neural network model as that in FedBN, with three convolutional layers with BN layers  and  three full-connected layers with BN layers. Please refer to Supplementary Materials \ref{SM::MS} for the details of this experiment.

Table \ref{tab::FedBN_results} illustrates that the FedDKD\_BN achieves the best test accuracy and outperforms  FedAvg by 3.21\%, FedProx by 3.14\%, FedBN by 0.31\%, and FedDKD by 0.58\%. Further,  FedDKD is  slightly weaker than FedBN in test accuracy. However, for FedBN, every local model maintains its BN layer, which means that it has $N$ models for $N$ clients, sharing most parameters and maintaining the respective batch normalization layer, whereas  FedDKD has only a single global model.

The DKD module can work with the model with BN layers. Moreover, combining the technique for FedBN can improve performance. This finding also implies that the DKD module can be a general tool that can be combined with other training methods for local models to improve performance.
\section{Conclusion}
This paper proposes a novel federated learning framework with decentralized knowledge distillation, called FedDKD.  Directly averaging the local models is not inherently the best method for the permutation invariance of neural networks and the heterogeneity of local datasets. Thus we  emphasize the importance of function space rather than parameter space. Furthermore,  FedDKD refines the initial averaging model according to the outputs of local models with KD. The global model is inspired to learn what local models learn, the local data distribution. Experiments on various datasets demonstrate that FedDKD can achieve superb test accuracy and even more efficient communication and faster convergence speeds in some extremely heterogeneous datasets.

In practice, FedDKD requires that the clients continuously  stay online several times, which is easy to implement in federated learning and does not increase any privacy risks compared to the universal algorithms. In FedDKD, local training is generally redundant, whereas communication is essential because  it helps determine the average in the function space rather than the parameter space for heterogeneous datasets.

\bibliographystyle{IEEEtran}
\bibliography{feddkd.bib}

\begin{thebibliography}{10}
\providecommand{\url}[1]{#1}
\csname url@samestyle\endcsname
\providecommand{\newblock}{\relax}
\providecommand{\bibinfo}[2]{#2}
\providecommand{\BIBentrySTDinterwordspacing}{\spaceskip=0pt\relax}
\providecommand{\BIBentryALTinterwordstretchfactor}{4}
\providecommand{\BIBentryALTinterwordspacing}{\spaceskip=\fontdimen2\font plus
\BIBentryALTinterwordstretchfactor\fontdimen3\font minus
  \fontdimen4\font\relax}
\providecommand{\BIBforeignlanguage}[2]{{%
\expandafter\ifx\csname l@#1\endcsname\relax
\typeout{** WARNING: IEEEtran.bst: No hyphenation pattern has been}%
\typeout{** loaded for the language `#1'. Using the pattern for}%
\typeout{** the default language instead.}%
\else
\language=\csname l@#1\endcsname
\fi
#2}}
\providecommand{\BIBdecl}{\relax}
\BIBdecl

\bibitem{FedAvg}
\BIBentryALTinterwordspacing
H.~B. McMahan, E.~Moore, D.~Ramage, S.~Hampson, and B.~A. y~Arcas,
  ``Communication-efficient learning of deep networks from decentralized
  data,'' in \emph{Proceedings of the 20th International Conference on
  Artificial Intelligence and Statistics (AISTATS)}, 2017. [Online]. Available:
  \url{http://arxiv.org/abs/1602.05629}
\BIBentrySTDinterwordspacing

\bibitem{FLLi}
\BIBentryALTinterwordspacing
T.~Li, A.~K. Sahu, A.~Talwalkar, and V.~Smith, ``Federated learning:
  Challenges, methods, and future directions,'' \emph{IEEE Signal Processing
  Magazine}, vol.~37, no.~3, p. 50–60, May 2020. [Online]. Available:
  \url{http://dx.doi.org/10.1109/MSP.2020.2975749}
\BIBentrySTDinterwordspacing

\bibitem{2015federated}
J.~Konečný, B.~McMahan, and D.~Ramage, ``Federated optimization:distributed
  optimization beyond the datacenter,'' 2015.

\bibitem{FedProx}
T.~Li, A.~K. Sahu, M.~Zaheer, M.~Sanjabi, A.~Talwalkar, and V.~Smith,
  ``Federated optimization in heterogeneous networks,'' 2020.

\bibitem{fedMAX}
W.~Chen, K.~Bhardwaj, and R.~Marculescu, ``Fedmax: Mitigating activation
  divergence for accurate and communication-efficient federated learning,'' in
  \emph{Machine Learning and Knowledge Discovery in Databases}, F.~Hutter,
  K.~Kersting, J.~Lijffijt, and I.~Valera, Eds.\hskip 1em plus 0.5em minus
  0.4em\relax Cham: Springer International Publishing, 2021, pp. 348--363.

\bibitem{FLNonIID}
\BIBentryALTinterwordspacing
Y.~Zhao, M.~Li, L.~Lai, N.~Suda, D.~Civin, and V.~Chandra, ``Federated learning
  with non-iid data.'' \emph{CoRR}, vol. abs/1806.00582, 2018. [Online].
  Available:
  \url{http://dblp.uni-trier.de/db/journals/corr/corr1806.html#abs-1806-00582}
\BIBentrySTDinterwordspacing

\bibitem{LargeScaleSGD}
\BIBentryALTinterwordspacing
L.~Bottou, ``Large-scale machine learning with stochastic gradient descent.''
  in \emph{COMPSTAT}, Y.~Lechevallier and G.~Saporta, Eds.\hskip 1em plus 0.5em
  minus 0.4em\relax Physica-Verlag, 2010, pp. 177--186. [Online]. Available:
  \url{http://dblp.uni-trier.de/db/conf/compstat/compstat2010.html#Bottou10}
\BIBentrySTDinterwordspacing

\bibitem{GD12}
A.~Rakhlin, O.~Shamir, and K.~Sridharan, ``Making gradient descent optimal for
  strongly convex stochastic optimization,'' 2012.

\bibitem{FedBN}
X.~Li, M.~Jiang, X.~Zhang, M.~Kamp, and Q.~Dou, ``Fedbn: Federated learning on
  non-iid features via local batch normalization,'' \emph{ArXiv}, vol.
  abs/2102.07623, 2021.

\bibitem{FedMA}
\BIBentryALTinterwordspacing
H.~Wang, M.~Yurochkin, Y.~Sun, D.~S. Papailiopoulos, and Y.~Khazaeni,
  ``Federated learning with matched averaging.'' in \emph{ICLR}.\hskip 1em plus
  0.5em minus 0.4em\relax OpenReview.net, 2020. [Online]. Available:
  \url{http://dblp.uni-trier.de/db/conf/iclr/iclr2020.html#WangYSPK20}
\BIBentrySTDinterwordspacing

\bibitem{EnsembleDistillation}
\BIBentryALTinterwordspacing
T.~Lin, L.~Kong, S.~U. Stich, and M.~Jaggi, ``Ensemble distillation for robust
  model fusion in federated learning.'' \emph{CoRR}, vol. abs/2006.07242, 2020.
  [Online]. Available:
  \url{http://dblp.uni-trier.de/db/journals/corr/corr2006.html#abs-2006-07242}
\BIBentrySTDinterwordspacing

\bibitem{AsynchronousOF}
Y.~Chen, Y.~Ning, M.~Slawski, and H.~Rangwala, ``Asynchronous online federated
  learning for edge devices with non-iid data,'' \emph{2020 IEEE International
  Conference on Big Data (Big Data)}, pp. 15--24, 2020.

\bibitem{2020Mix2FLD}
S.~Oh, J.~Park, E.~Jeong, H.~Kim, and S.~L. Kim, ``Mix2fld: Downlink federated
  learning after uplink federated distillation with two-way mixup,'' \emph{IEEE
  Communications Letters}, vol.~PP, no.~99, pp. 1--1, 2020.

\bibitem{2020FedED}
D.~Sui, Y.~Chen, J.~Zhao, Y.~Jia, and W.~Sun, ``Feded: Federated learning via
  ensemble distillation for medical relation extraction,'' in \emph{Proceedings
  of the 2020 Conference on Empirical Methods in Natural Language Processing
  (EMNLP)}, 2020.

\bibitem{fedistill}
H.~Seo, J.~Park, S.~Oh, M.~Bennis, and S.-L. Kim, ``Federated knowledge
  distillation,'' 2020.

\bibitem{2021FedGen}
Z.~Zhu, J.~Hong, and J.~Zhou, ``Data-free knowledge distillation for
  heterogeneous federated learning,'' 2021.

\bibitem{FedNTD}
\BIBentryALTinterwordspacing
G.~Lee, M.~Jeong, Y.~Shin, S.~Bae, and S.-Y. Yun, ``Preservation of global
  knowledge by not-true distillation in federated learning,'' 2021. [Online].
  Available: \url{https://arxiv.org/abs/2106.03097}
\BIBentrySTDinterwordspacing

\bibitem{FedGDK}
\BIBentryALTinterwordspacing
D.~Yao, W.~Pan, Y.~Dai, Y.~Wan, X.~Ding, H.~Jin, Z.~Xu, and L.~Sun,
  ``Local-global knowledge distillation in heterogeneous federated learning
  with non-iid data,'' 2021. [Online]. Available:
  \url{https://arxiv.org/abs/2107.00051}
\BIBentrySTDinterwordspacing

\bibitem{2020Astraea}
\BIBentryALTinterwordspacing
M.~Duan, ``Astraea: Self-balancing federated learning for improving
  classification accuracy of mobile deep learning applications.'' \emph{CoRR},
  vol. abs/1907.01132, 2019. [Online]. Available:
  \url{http://dblp.uni-trier.de/db/journals/corr/corr1907.html#abs-1907-01132}
\BIBentrySTDinterwordspacing

\bibitem{1951KL}
S.~Kullback and R.~A. Leibler, ``On information and sufficiency,'' \emph{Ann.
  Math. Statist.}, vol.~22, no.~1, pp. 79--86, 1951.

\bibitem{books/daglib/0035708}
A.~B. Tsybakov, \emph{Introduction to Nonparametric Estimation.}, ser. Springer
  series in statistics.\hskip 1em plus 0.5em minus 0.4em\relax Springer, 2009.

\bibitem{FEMNIST}
\BIBentryALTinterwordspacing
S.~Caldas, P.~Wu, T.~Li, J.~Konecný, H.~B. McMahan, V.~Smith, and
  A.~Talwalkar, ``Leaf: A benchmark for federated settings.'' \emph{CoRR}, vol.
  abs/1812.01097, 2018. [Online]. Available:
  \url{http://dblp.uni-trier.de/db/journals/corr/corr1812.html#abs-1812-01097}
\BIBentrySTDinterwordspacing

\bibitem{cifar-10}
\BIBentryALTinterwordspacing
A.~Krizhevsky, ``Learning multiple layers of features from tiny images,'' pp.
  32--33, 2009. [Online]. Available:
  \url{https://www.cs.toronto.edu/~kriz/learning-features-2009-TR.pdf}
\BIBentrySTDinterwordspacing

\bibitem{SVHN}
\BIBentryALTinterwordspacing
Y.~Netzer, T.~Wang, A.~Coates, A.~Bissacco, B.~Wu, and A.~Y. Ng, ``Reading
  digits in natural images with unsupervised feature learning,'' 2011.
  [Online]. Available:
  \url{http://ufldl.stanford.edu/housenumbers/nips2011_housenumbers.pdf}
\BIBentrySTDinterwordspacing

\bibitem{USPS}
\BIBentryALTinterwordspacing
J.~J. Hull, ``A database for handwritten text recognition research.''
  \emph{IEEE Trans. Pattern Anal. Mach. Intell.}, vol.~16, no.~5, pp. 550--554,
  1994. [Online]. Available:
  \url{http://dblp.uni-trier.de/db/journals/pami/pami16.html#Hull94}
\BIBentrySTDinterwordspacing

\bibitem{SynthDigits-MNIST-M}
\BIBentryALTinterwordspacing
Y.~Ganin and V.~S. Lempitsky, ``Unsupervised domain adaptation by
  backpropagation.'' in \emph{ICML}, ser. JMLR Workshop and Conference
  Proceedings, F.~R. Bach and D.~M. Blei, Eds., vol.~37.\hskip 1em plus 0.5em
  minus 0.4em\relax JMLR.org, 2015, pp. 1180--1189. [Online]. Available:
  \url{http://dblp.uni-trier.de/db/conf/icml/icml2015.html#GaninL15}
\BIBentrySTDinterwordspacing

\bibitem{MNIST}
Y.~LeCun, L.~Bottou, Y.~Bengio, and P.~Haffner, ``Gradient-based learning
  applied to document recognition,'' \emph{Proceedings of the IEEE}, vol.~86,
  no.~11, pp. 2278--2324, 1998.

\end{thebibliography}

\vfill
\clearpage

\appendix
\section*{Supplementary Materials}
\par
\subsection{The proof of Proposition \ref{prop::1}} \label{SM::TVD}
\begin{proof}
By the triangular inequality of divergence $\delta$, we have
\begin{equation*}
\begin{aligned}
& \int \delta^\dag(\Phi^*(x), \Phi(x;w))P(dx)\\
& \leq \int [\delta(\Phi^*(x), \mathcal{T}(x))+ \delta(\mathcal{T}(x), \Phi(x;w))]^2 P(dx) \\
\end{aligned}
\end{equation*}
for ${\forall}$ probability distribution $\mathcal{T}(x)$.
\par
The total variation distance satisfies the Pinsker’s inequality:
\begin{equation*}
\begin{aligned}
& \delta(P(x), Q(x))\leq \sqrt{\frac{1}{2}D_{KL}(P(x)||Q(x))}
\end{aligned}
\end{equation*}
Then, we have the following:
$$
	\begin{aligned}
	 & \int \delta^\dag(\Phi^*(x), \Phi(x;w))P(dx) \\
	 & \leq \int [\delta(\Phi^*(x), \mathcal{T}(x))+ \delta(\mathcal{T}(x), \Phi(x;w))]^2 P(dx) \\
	 & \leq \int\{\sqrt{\frac{D_{KL}(\Phi^*(x)||\mathcal{T}(x))}{2}} \\
	 & \hspace{0.5cm}+ \sqrt{\frac{D_{KL}(\mathcal{T}(x)|| \Phi(x;w))}{2}}\}^2P(dx)
	 \\
	& = \int \{\frac{D_{KL}(\Phi^*(x)||\mathcal{T}(x))}{2} +  \frac{D_{KL}(\mathcal{T}(x)|| \Phi(x;w))}{2} \\
	& \hspace{0.5cm}+ \sqrt{D_{KL}(\Phi^*(x)||\mathcal{T}(x)) \cdot D_{KL}(\mathcal{T}(x)|| \Phi(x;w))} \}P(dx)\\
	&  \leq \sum_{k=1}^K q_k \int_{\mathcal D_{k}} D_{KL}(\Phi^*(x)|| \Phi(x;w_k))P_k(dx) \\
    & \hspace{0.5cm}+ \sum_{k=1}^K q_k \int_{\mathcal D_{k}} D_{KL}(\Phi(x;w_k)|| \Phi(x;w))P_k(dx)
	\end{aligned}
$$
where $\mathcal{T}(x) = \sum_{k=1}^K \Phi(x;w_k) \mathbbm{1}_{\mathcal D_{k}}(x)$.

 \end{proof}
\subsection{Analysis of the DKD Module} \label{SM::proof_theorem1}
\textbf{The proof of Proposition \ref{prop::2}}
\begin{proof}
As we defined in the main body, the second-order Taylor approximation of the optimized function in Eq. (\ref{equ::dkd}) on a fixed point $w^*$ is as follows:
$$
	\begin{aligned}
		& \sum_{k=1}^{K} q_k \int_{\mathcal{D}_k}div(\Phi(x; w_k), \Phi(x; w))P_k(dx) \\
		= & \sum_{k=1}^{K} q_k \int_{\mathcal{D}_k}div(\Phi(x; w^* + (w_k - w^*)), \\
		&        \hspace{1cm}  \Phi(x; w^* + (w - w^*)))P_k(dx) \\
		= & \frac{1}{2}\sum_{k=1}^{K} q_k \langle w_k - w\vert \int_{\mathcal{D}_k}\frac{\partial \Phi(x; w^*)}{\partial w^*}^{\top} \\
		& \hspace{1cm} \cdot H_{\Phi}(x; w^*)\frac{\partial \Phi(x; w^*)}{\partial w^*} P_k(dx)\vert w_k - w\rangle \\
		 & \hspace{1cm} + O(\|w_k - w^*\|^3) + O(\|w - w^*\|^3)\\
		:= & \frac{1}{2}\sum_{k=1}^{K} q_k \langle w_k - w\vert H_k(w^*) \vert w_k - w \rangle \\
		&\hspace{1cm}+ O(\|w_k - w^*\|^3) + O(\|w - w^*\|^3)
	\end{aligned}
$$
	Moreover, $H(w^*)$ is at least a semi-positive matrix. Hence, $\frac{1}{2}\sum_{k=1}^{K} q_k \langle w_k - w\vert H_k(w^*) \vert w_k - w \rangle$ is convex for $w$. We only need to verify that $\sum_{k=1}^{K}q_kw_k$ is one of its stationary points, which is obvious.
\end{proof}

\textbf{The proof of Corollary \ref{cor::1}}
\begin{proof}
    This corollary is a linear form of Corollary \ref{cor::1}. The choice of the fixed point $w^*$ in the Taylor approximation is irrelevant. Hence, the effect of different values of $w^*$   in Eq. (\ref{approx}) is only included in $O(\|w_k - w^*\|^3) + O(\|w - w^*\|^3)$. However, this term does not exist for the linear model, and the Taylor approximation is precise, proving the correctness of this corollary.
\end{proof}


\subsection{Pseudocode of the local training schema of FedDKD\_MAX}\label{SM::FedDKD_fedMAX}
Algorithm \ref{algorithm::FedDKD_fedMAX} provides the pseudocode of the local training schema of FedDKD\_MAX where $F(w;b)$ is the cross-entropy loss, $\|\cdot\|$ is the $L^2$ norm, and $a_i$ refers to the activation vectors at the input of the last fully-connected layer of sample $i$.
\begin{algorithm}[tb]  
\caption{The Client-LocalTrain of \textit{FedDKD\_MAX}}
\label{algorithm::FedDKD_fedMAX}
\textbf{Parameters}: $\beta$. \\
\textbf{Client-LocalTrain}$(k,w)$:
\begin{algorithmic}[1]
\STATE
$\mathcal{B} \longleftarrow$ split $\mathcal{D}_k$ into batches of size $B$
\FOR{\text{each local epoch} $i=1,2,\cdots E$}
\FOR {batch $b$ in $\mathcal{D}_k$}
\STATE
$g(w; b)=F(w;b) + \beta \frac{1}{B} \sum_{i=1}^B KL(a_i\|U)$; \\
\STATE
$w \longleftarrow w - \eta \nabla g(w;b)$;
\ENDFOR
\ENDFOR
\STATE
\textbf{return} $w$ to server
\end{algorithmic}
\end{algorithm}

\subsection{Pseudocode of FedDKD\_BN} \label{SM::FedDKD_FedBN}
The FedDKD\_BN algorithm is described as Algorithm \ref{algorithm::FedDKD_FedBN}, and the DKDSGD is modified as Algorithm \ref{algorithm::DKDSGD_feddkd_fedbn}.
\begin{algorithm}[h] 
\caption{\textit{DKDSGD\_BN} for FedDKD\_BN. Solver with SGD on each client $k$ in step $j$}
\label{algorithm::DKDSGD_feddkd_fedbn}
\textbf{Input}:The global parameters $w'$ (no BN layers), the parameters of client $k$'s model $w_k$, the parameters of client $k$'s BN layers $w^{BN}_k$, local training dataset $\mathcal{D}_k$; \\
\textbf{Parameter}: DKD mini-batch size $\Tilde{B}$; \\
\textbf{Output}:The local gradient $\nabla d_{k,j}$.

\begin{algorithmic}[1]
\STATE
Sample a min-batch $\mathcal{B}_k$\ from training data $\mathcal{D}_k$ with batch size $\Tilde{B}$ ;
\STATE
Get the logit of the $k$st local model $\Phi_k(\mathcal{B}_k, w_k)$ and the logit of the global model  $\Phi(\mathcal{B}_k; w', w^{BN}_k)$;
\STATE
$\nabla d_{k,j} = \frac{\partial 
L_{CE}(\Phi_k(\mathcal{B}_k, w_k), \Phi(\mathcal{B}_k; w', w^{BN}_k))
}{\partial w'}$
\STATE
\textbf{return} $\nabla d_{k,j}$
\end{algorithmic}
\end{algorithm}

\begin{algorithm}[htb] 
\caption{\textit{FedDKD\_BN}}
\label{algorithm::FedDKD_FedBN}
\textbf{Input}:Private datasets $\mathcal{D}_k, k=1,2,\cdots, K$ \\
\textbf{Federated learning parameters}:the local epochs $E$, local mini-batch size $B$, the fraction of clients involved  each round $C$, local learning rate $\eta$, the DKD rounds $T$ \\
\textbf{DKD parameters}:
 DKD steps $J$,  DKD learning rate $\gamma$, DKD mini-batch size $\Tilde{B}$ \\
\textbf{Output}: The updated global model $w^*_T$ \\
\textbf{procedure} SERVER ; \par
\begin{algorithmic}[1]
\FOR{Each DKD round $t=1,\cdots, T$}
\STATE
$m \longleftarrow  \max(C\cdot K, 1)$
\STATE
Randomly sample a set of $m$ clients $\mathcal{C}_t$
\STATE
For each client $k\in \mathcal{C}_i$, in parallel do
\STATE
$w_k$ $\longleftarrow$ Client-LocalTrain$(k, w^{*}_{t-1})$
\STATE
Get the initial global parameter \\
\FOR {each layer $l$}
\IF{layer $l$ is not BN layer}
\STATE $\bar{w}'^{(l)}_{t,0}=\sum_{k\in \mathcal{C}_t} \frac{n_k}{\sum_{i\in \mathcal{C}_t} n_i} w^{(l)}_k$
\ENDIF
\ENDFOR
\FOR{$j\in \{1,\cdots, J\}$}
\STATE
For each client $k\in \mathcal{C}_i$, in parallel do
\STATE
        $\nabla d_{k,j}$ $\longleftarrow$ \textit{DKDSGD\_BN}$(\bar{w}'_{t,j-1}, w_k, w^{BN}_k, \mathcal{D}_k)$
\STATE
        $\bar{w}'_{t,j} = \bar{w}'_{t,j-1} - \gamma (\frac{1}{m}\sum_{k \in \mathcal{C}_t} \nabla d_{k,j})$
\ENDFOR
\STATE
$w^{*}_t=\{\bar{w'}_{t, J}, w^{BN}_k, k\in \mathcal{C}_t \}$
\ENDFOR
\STATE
\textbf{return} $w_T^*$
\end{algorithmic}
\textbf{Client-LocalTrain}$(k,w', w^{BN})$:
\begin{algorithmic}[1]
\STATE
$\mathcal{B} \longleftarrow$ split $\mathcal{D}_k$ into batches of size $B$
\FOR{\text{each local epoch} $i=1,2,\cdots E$}
\FOR {batch $b$ in $\mathcal{D}_k$}
\STATE
$w \longleftarrow \text{optimizer}(b;w',w^{BN};\eta)$
\ENDFOR
\ENDFOR
\STATE
\textbf{return} $w$ to server
\end{algorithmic}
\end{algorithm}
\subsection{Experiments on FEMNIST}\label{SM::FEMNIST}
\textbf{Model Architecture}.
For the experiments on the FEMNIST dataset, we use a "slimmer" CNN model than that in \cite{fedMAX}, removing the last two convolutional layers. The employed CNN model has three convolutional layers whose channels are 32, 64, and 64
 and two full-connected layers. A rectified linear unit (ReLU) activation layer and max-pooling layer follow each convolutional layer. A ReLU layer activates the output of the first fully connected layer, and the last one yields the output for the model. The details are listed in Table \ref{tab::femnist}.
\begin{table}[H]
    \centering
        \caption{Model architecture for FEMNIST dataset. }
    \begin{tabular}{c|c }
    \hline
         Layer & parameters \\
         \hline
         1 & Conv2d(1, 32),  ReLU, MaxPool2D \\
         \hline
         2 & Conv2d(32, 64),  ReLU, MaxPool2D \\
         \hline
         3 & Conv2d(64, 64),  ReLU, MaxPool2D \\
         \hline
         4 &  FC(256, 512), ReLU \\
         \hline
         5 & FC(512, 26) \\
         \hline
    \end{tabular}
    \label{tab::femnist}
\end{table}

\textbf{Detailed Settings and Hyper-parameter Searching}.
On local training in the clients, the learning rate $\eta$ is initialized as $0.1$, and the decay rate is $0.98$ per DKD round. The batch size $B$ is $64$. The optimizer is the SGD with momentum, and the weight decay rate is $1e-3$ for $c=0.25$ where $E=5$, and otherwise, is $5e-4$.

For FedProx, we perform a grid search with $\mu\in\{0.01, 0.1, 1, 10\}$. We  choose the best one  ($10$). For FedMAX, we test $\beta\in\{1, 10, 100, 1000, 10000\}$, and the model with $\beta = 10$ achieves the best performance. For FedDKD and FedDKD\_MAX, we set the DKD learning rate $\gamma=0.2$ with the same decay rate of $0.98$, with DKD steps $J=10$ and a DKD batch size $\Tilde{B}$ of $64$.

\begin{figure}
    \centering
    \includegraphics[width=0.48\columnwidth]{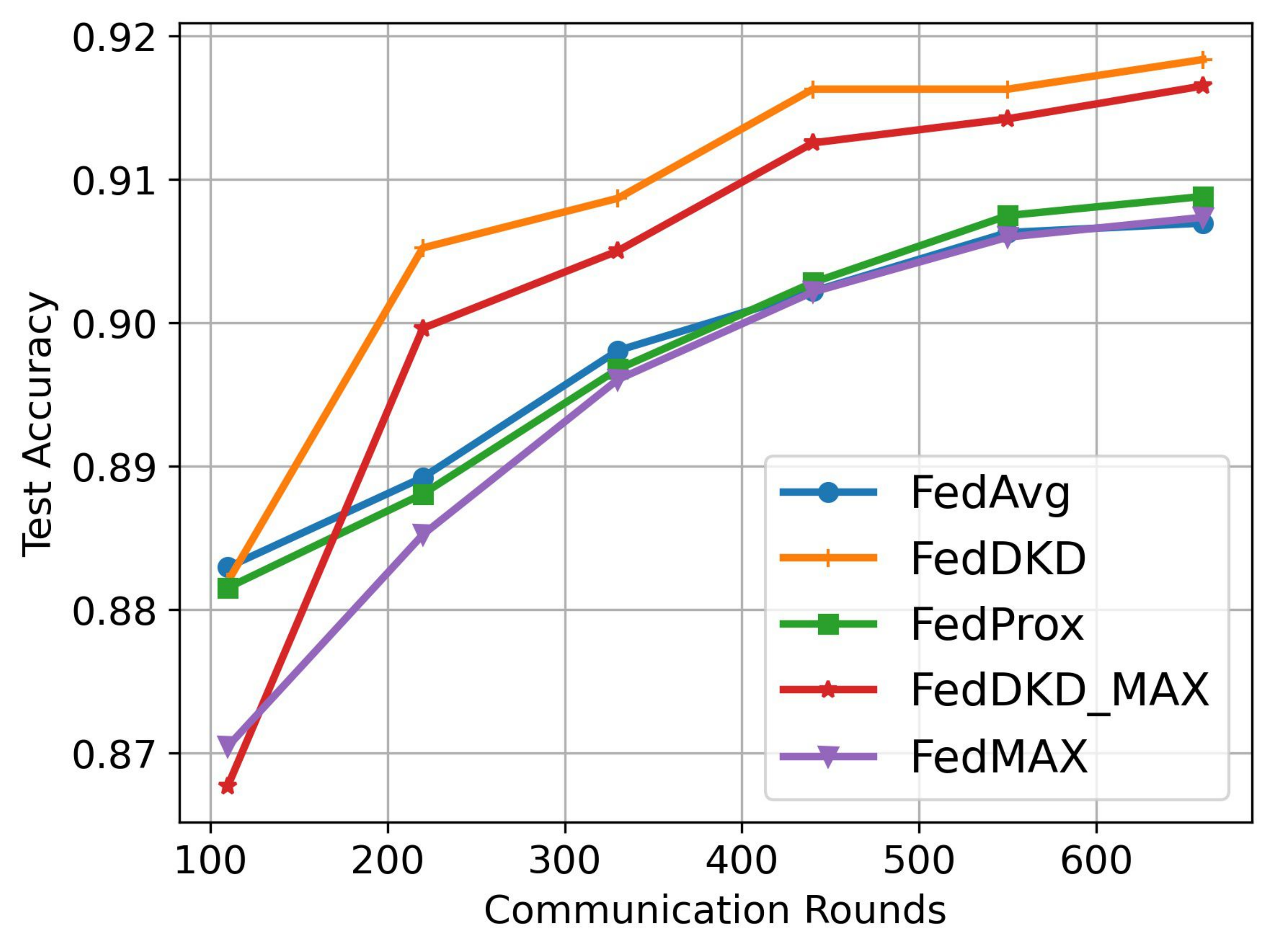}
    \includegraphics[width=0.48\columnwidth]{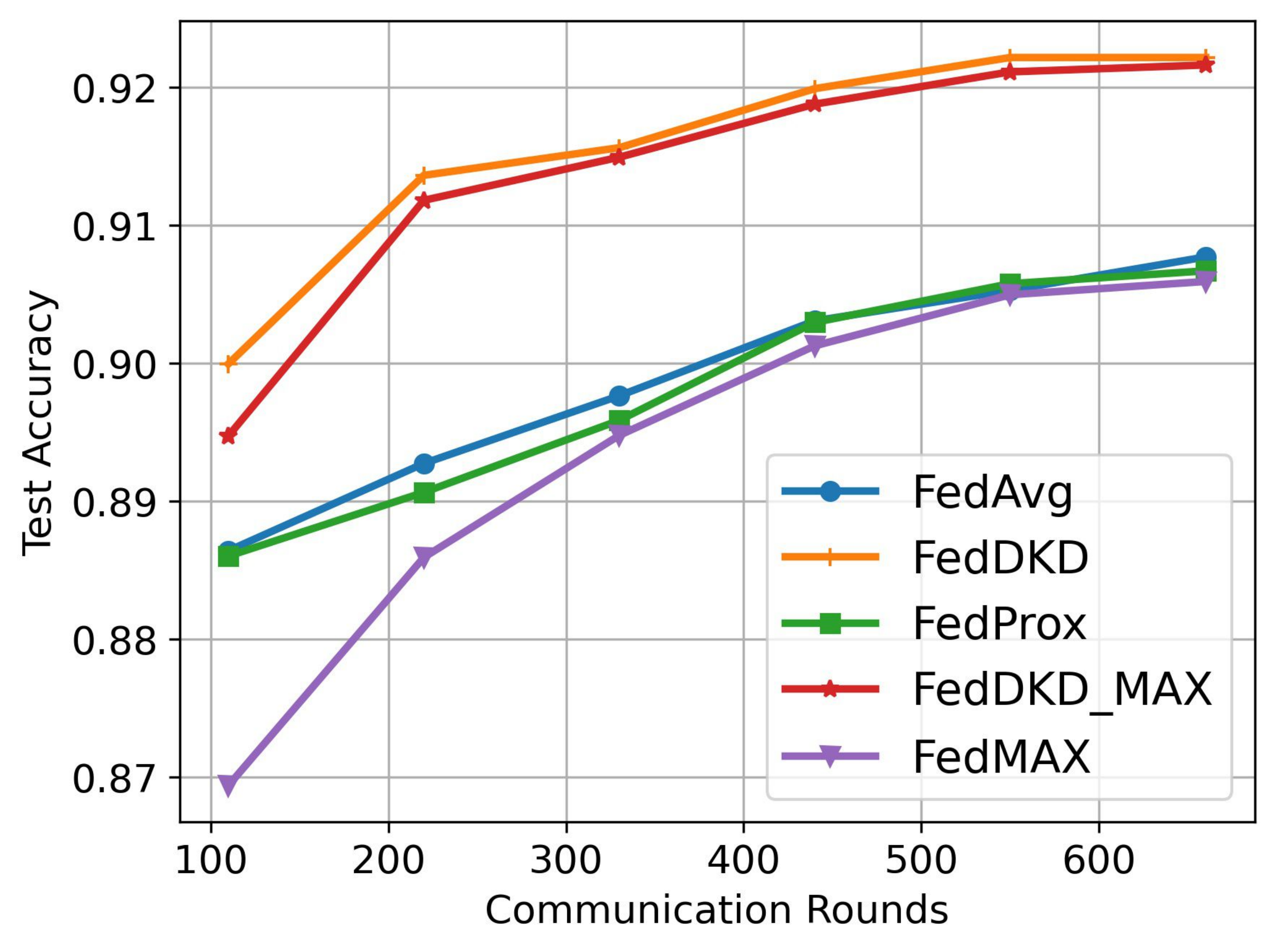}
    \caption{Test accuracy under the same communication cost for $c=0.5$ (left) and $c=0.75$ (right), and $E=15$.}
    \label{fig::E15_}
\end{figure}

\textbf{Communication-Efficient for $E=15$.} We further show our method is communication-efficient when $E=15$ in Figure \ref{fig::E15_}.

\begin{figure}
    \centering
    \includegraphics[width=0.48\columnwidth]{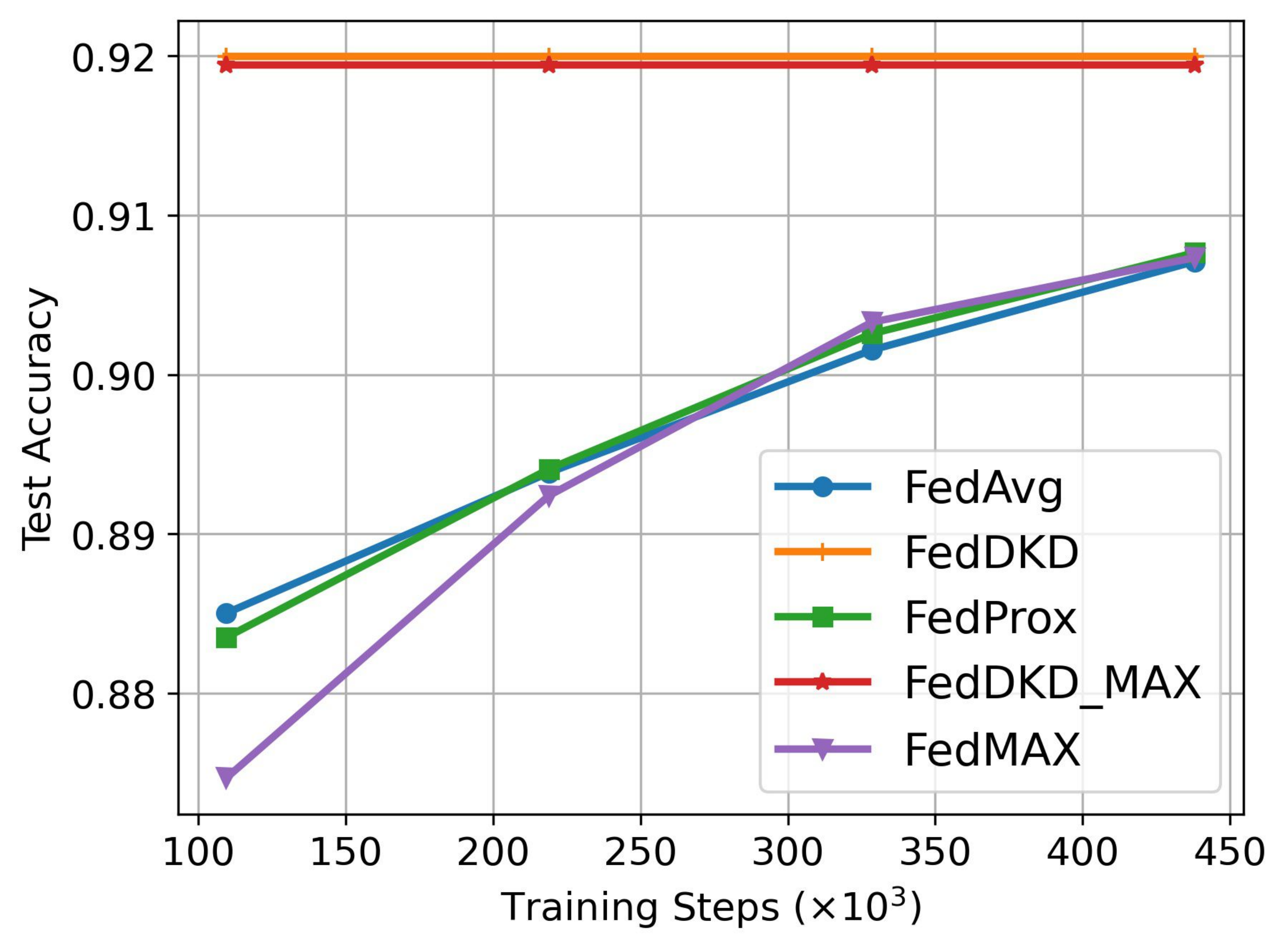}
    \includegraphics[width=0.48\columnwidth]{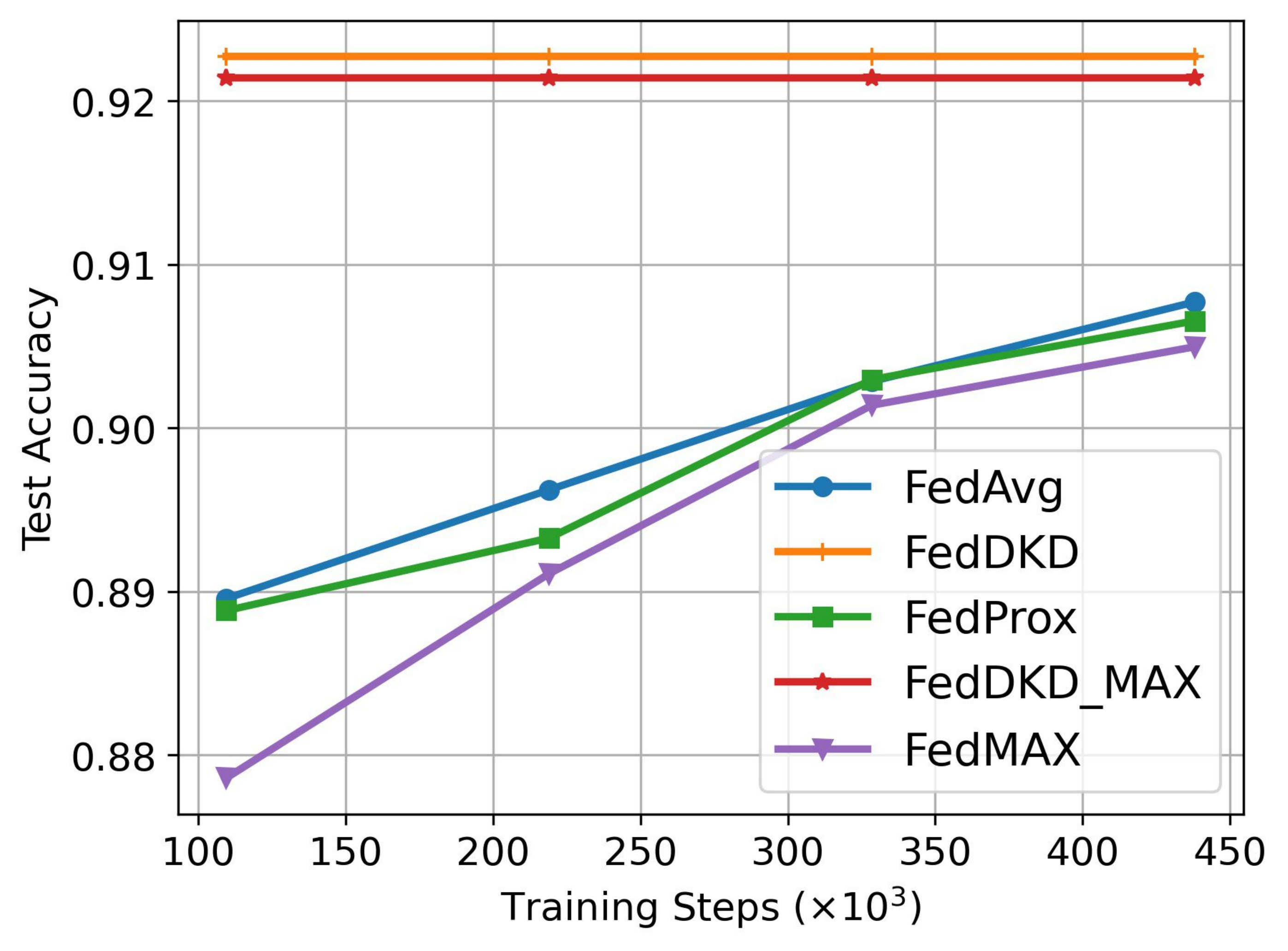}
    \caption{Test accuracy under the same training steps for $c=0.5$ (left) and $c=0.75$ (right) when $E=15$.}
    \label{fig::E15_step}
\end{figure}

\begin{figure}
    \centering
    \includegraphics[width=0.48\columnwidth]{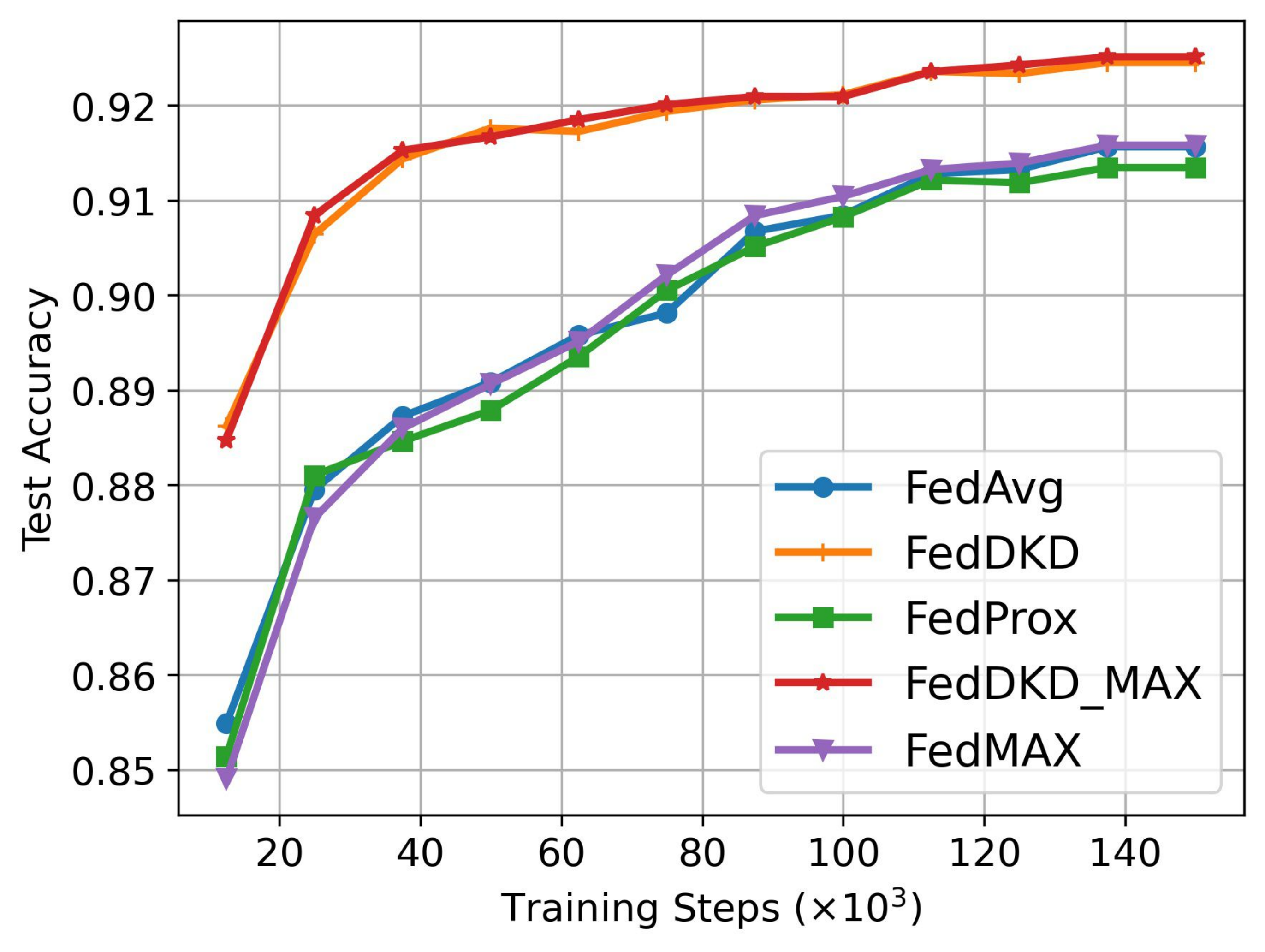}
    \includegraphics[width=0.48\columnwidth]{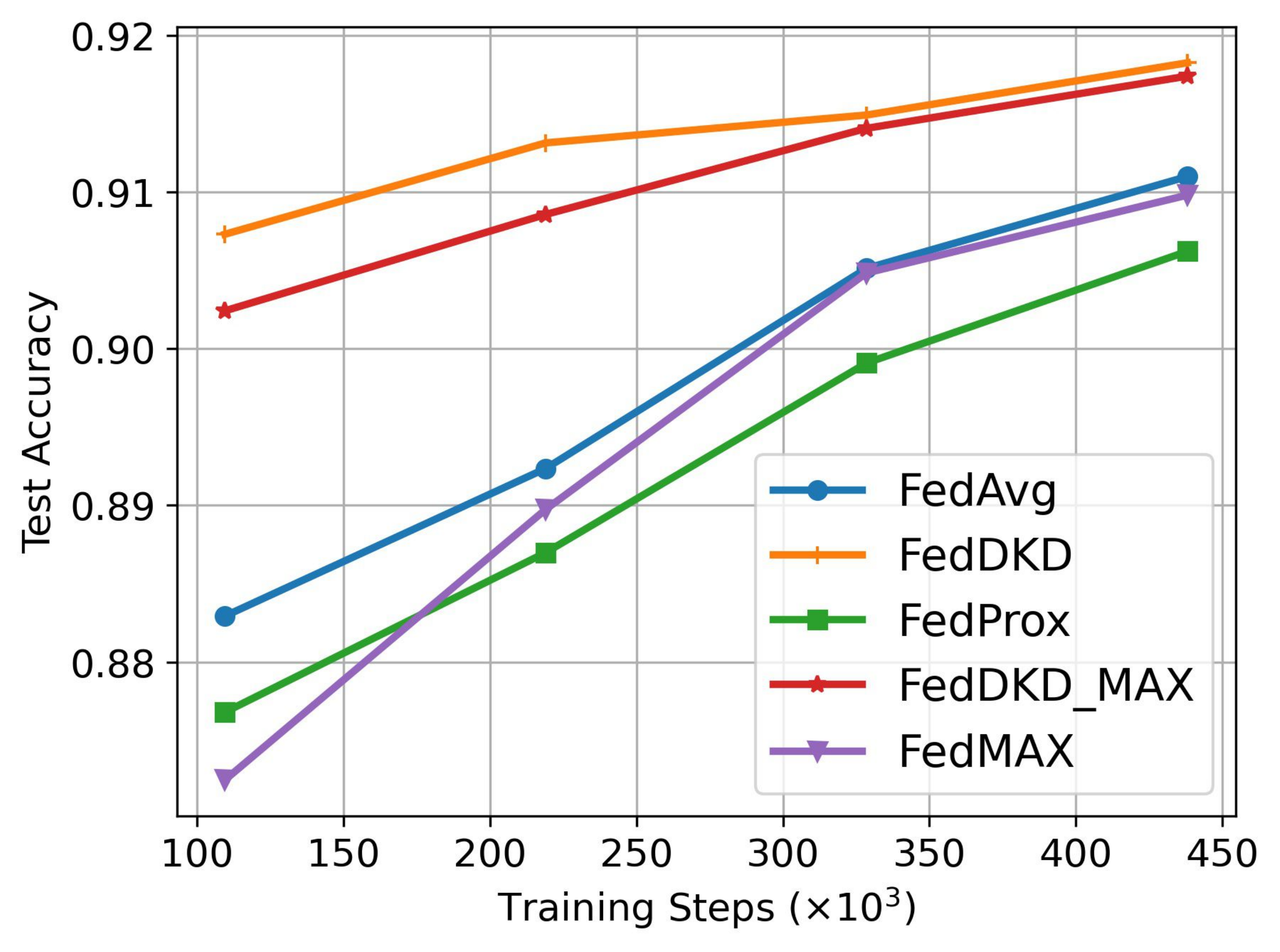}
    \caption{Test accuracy under the same training steps for $E=5$ (left) and $E=15$ (right) when $c=0.25$.}
    \label{fig::c0.25_}
\end{figure}

\textbf{Training-Efficient for other settings}. We further show our method is Training-efficient when $c=0.25$ in Figure \ref{fig::E15_step} and \ref{fig::c0.25_}.

\subsection{Experiments on CIFAR-10/100}\label{SM::CIFAR10}
\textbf{Model Architecture}
For the experiments on the CIFAR-10/100 dataset, we use the same CNN model as that in \cite{FedMA} (i.e., VGG-9), and the details are listed in Table \ref{tab::cifar_vgg9}.
\begin{table}[htp]
    \centering
    \caption{Model architecture for the CIFAR-10/100 dataset. }
    \begin{tabular}{c|c }
    \hline
         Layer & parameters \\
         \hline
         1 & Conv2d(3, 32),  ReLU \\
         \hline
         2 & Conv2d(32, 64),  ReLU, MaxPool2D \\
         \hline
         3 & Conv2d(64, 128),  ReLU \\
         \hline
         4 & Conv2d(128, 128),  ReLU, MaxPool2D \\
         \hline
         5 & Dropout2d(0.05) \\
         \hline
         6 &  Conv2d(128, 256),  ReLU \\
         \hline
         7 & Conv2d(256, 256),  ReLU, MaxPool2D \\
         \hline
         8 & Dropout(0.1) \\
         \hline
         9 &  FC(4096, 512), ReLU \\
         \hline
         10 & FC(512, 512), ReLU \\
         \hline
         11 & Dropout(0.1) \\
         \hline
         12 & FC(512, 10/100) \\
         \hline
    \end{tabular}
    \label{tab::cifar_vgg9}
\end{table}

\textbf{Detailed Settings}
On local training of the experiments on CIFAR-10/100, the number of epochs $E$ is $10$, and the batch size $B$ is $64$. The optimizer is Adam, with a learning rate of $1e-3$ and a weight decay rate of $1e-4$. Moreover, the learning rate has a decay rate of $0.99$ per DKD round. The total number of DKD rounds is $350$.

 For FedProx, we search the hyper-parameter $\mu\in\{0.001, 0.01, 0,1, 1, 10\}$ and choose the best one (0.01) for both CIFAR-10 and CIFAR-100.

For other FedDKD settings, we set DKD batch size $\Tilde{B}$ to $64$ for CIFAR-10, which equals the local training batch size $B$. For CIFAR-100, we enlarge the DKD batch size $\Tilde{B}$ to $256$ beacause the total number of categories is $100$.
\par

\subsection{Experiments on Multi-sources digits datasets} \label{SM::MS}
\textbf{Model Architecture}.
For the experiments on multi-sources digits datasets, we use the same CNN model as that in FedBN. The details are listed in Table \ref{tab::fedbn_model}.
\begin{table}[H]
    \centering
        \caption{Model architecture for Multi-sources digits datasets. }
    \begin{tabular}{c|c }
    \hline
         Layer & parameters \\
         \hline
         1 & Conv2d(3, 64), BN(64), ReLU, MaxPool2D \\
         \hline
         2 & Conv2d(64, 64), BN(64), ReLU, MaxPool2D \\
         \hline
         3 & Conv2d(64, 128), BN(128), ReLU \\
         \hline
         4 &  FC(6272, 2048), BN(2048), ReLU \\
         \hline
         5 &  FC(2048, 512), BN(512), ReLU \\
         \hline
         6 & FC(512, 10) \\
         \hline
    \end{tabular}
    \label{tab::fedbn_model}
\end{table}

\textbf{Detailed Settings}.
For the federated learning settings, the number of clients is $5$, and all clients participate in the integration per round. For local training, the number of epochs is $1$, and the batch size is 32. The optimizer is SGD with a learning rate of $0.01$ and a decay rate of $0.98$. The total DKD round is 100 in this paper.

For the FedDKD, the DKD learning rate is $0.01$, with a deacy rate of $0.98$. The number of DKD steps is $10$, and the DKD batch size is 10. In each DKD round, the DKD learning rate also has a decay rate of $0.97$ per DKD step.

\begin{figure}[h]
    \centering
    \includegraphics[width=0.45\columnwidth]{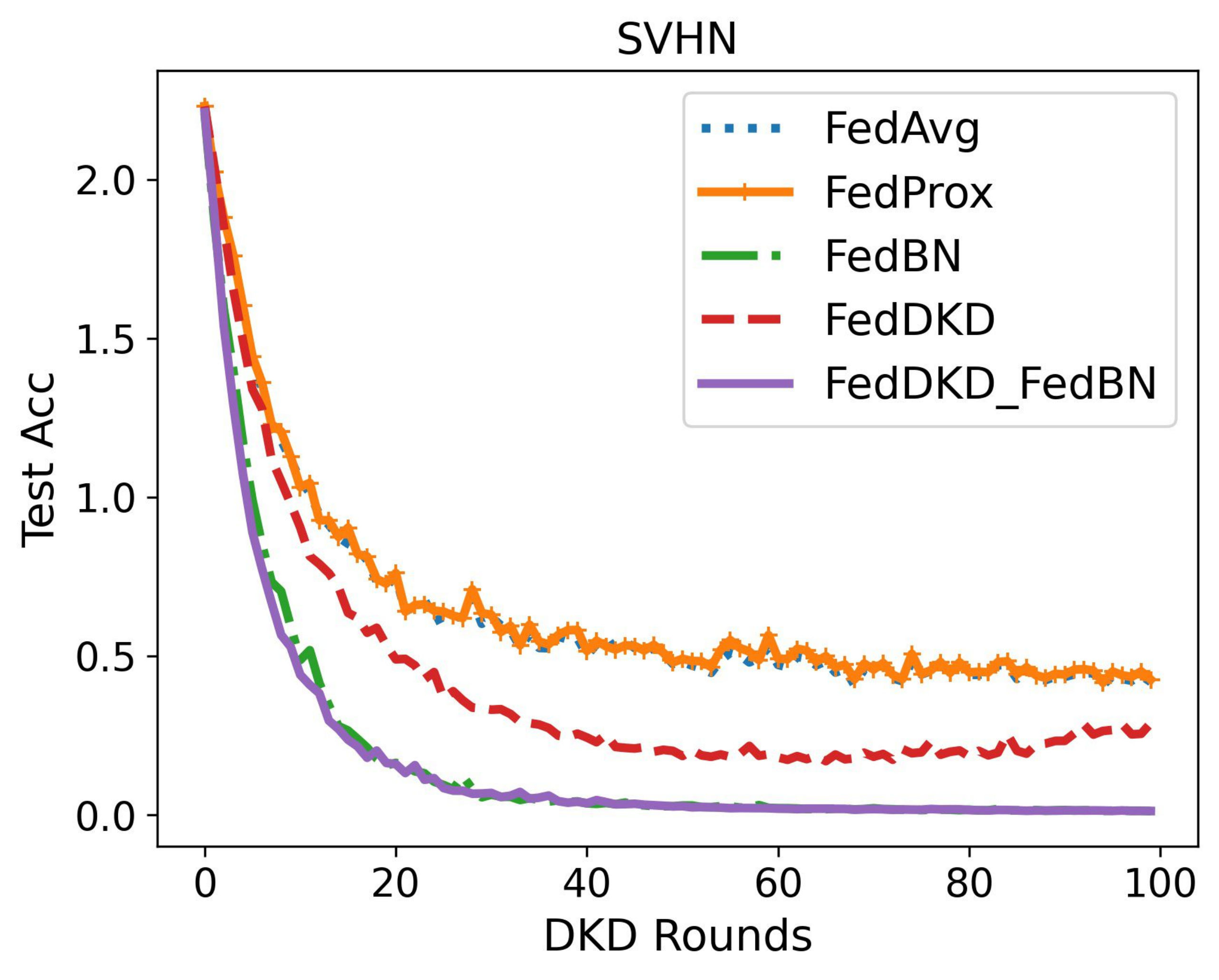}
    \includegraphics[width=0.45\columnwidth]{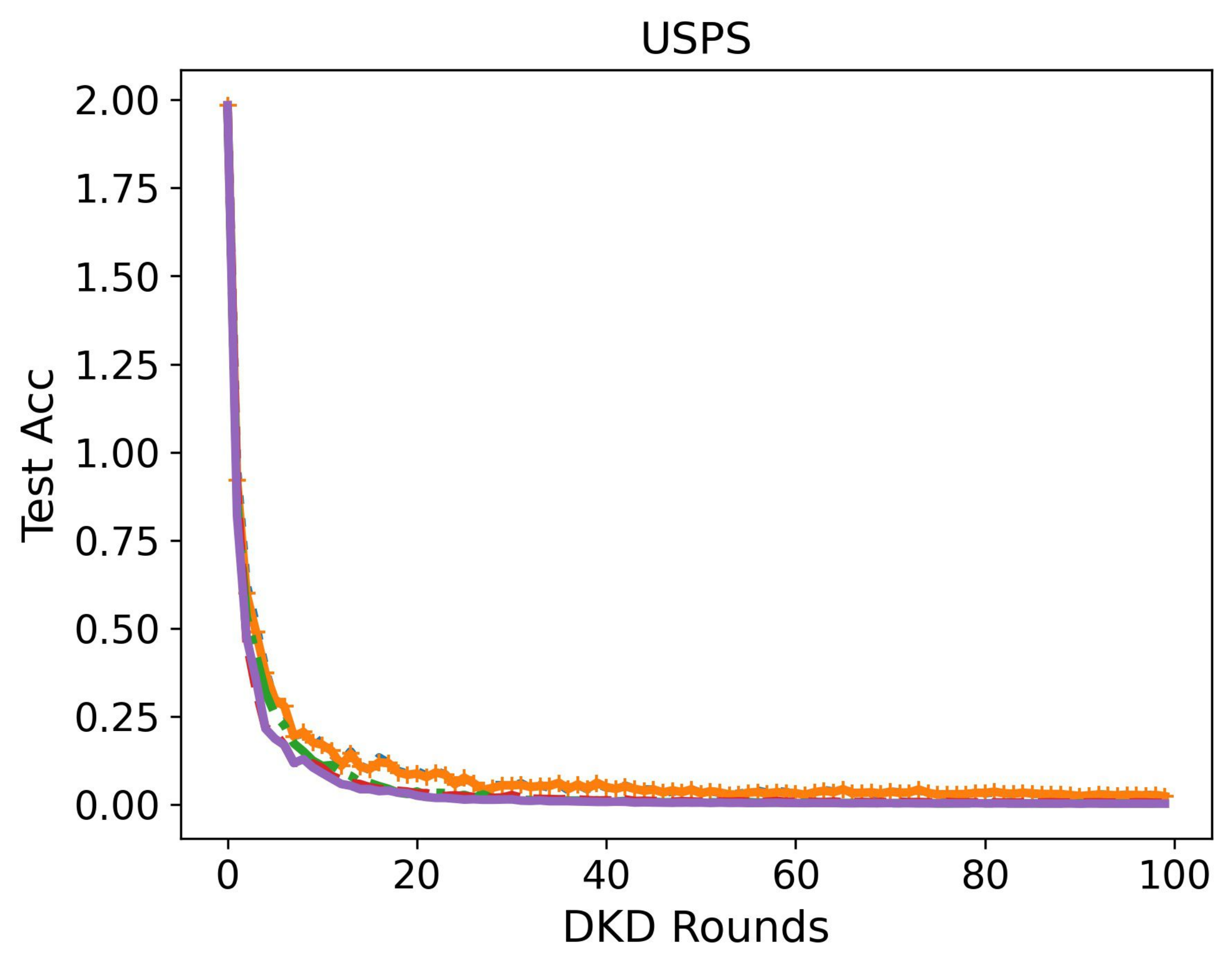}
    \includegraphics[width=0.45\columnwidth]{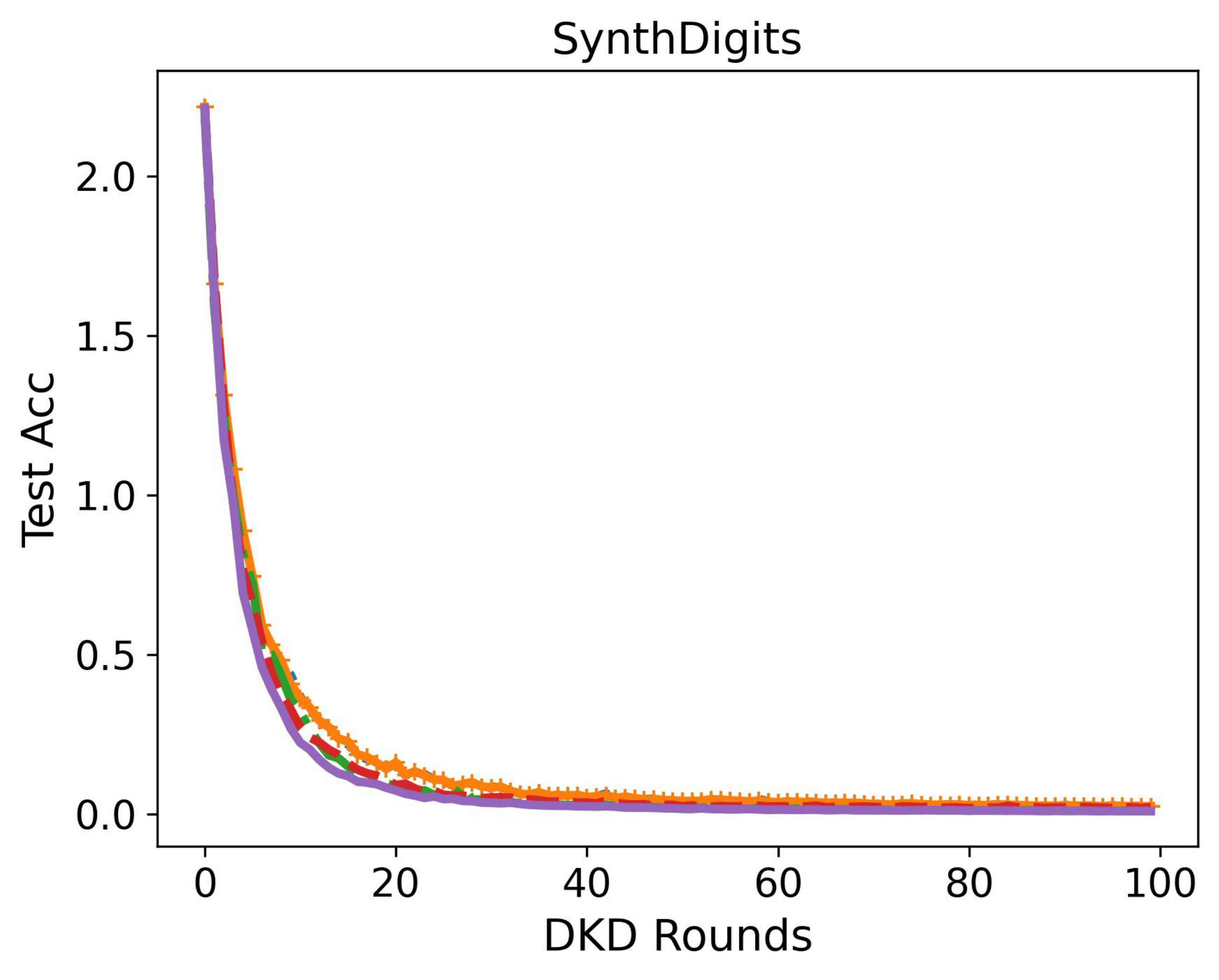}
    \includegraphics[width=0.45\columnwidth]{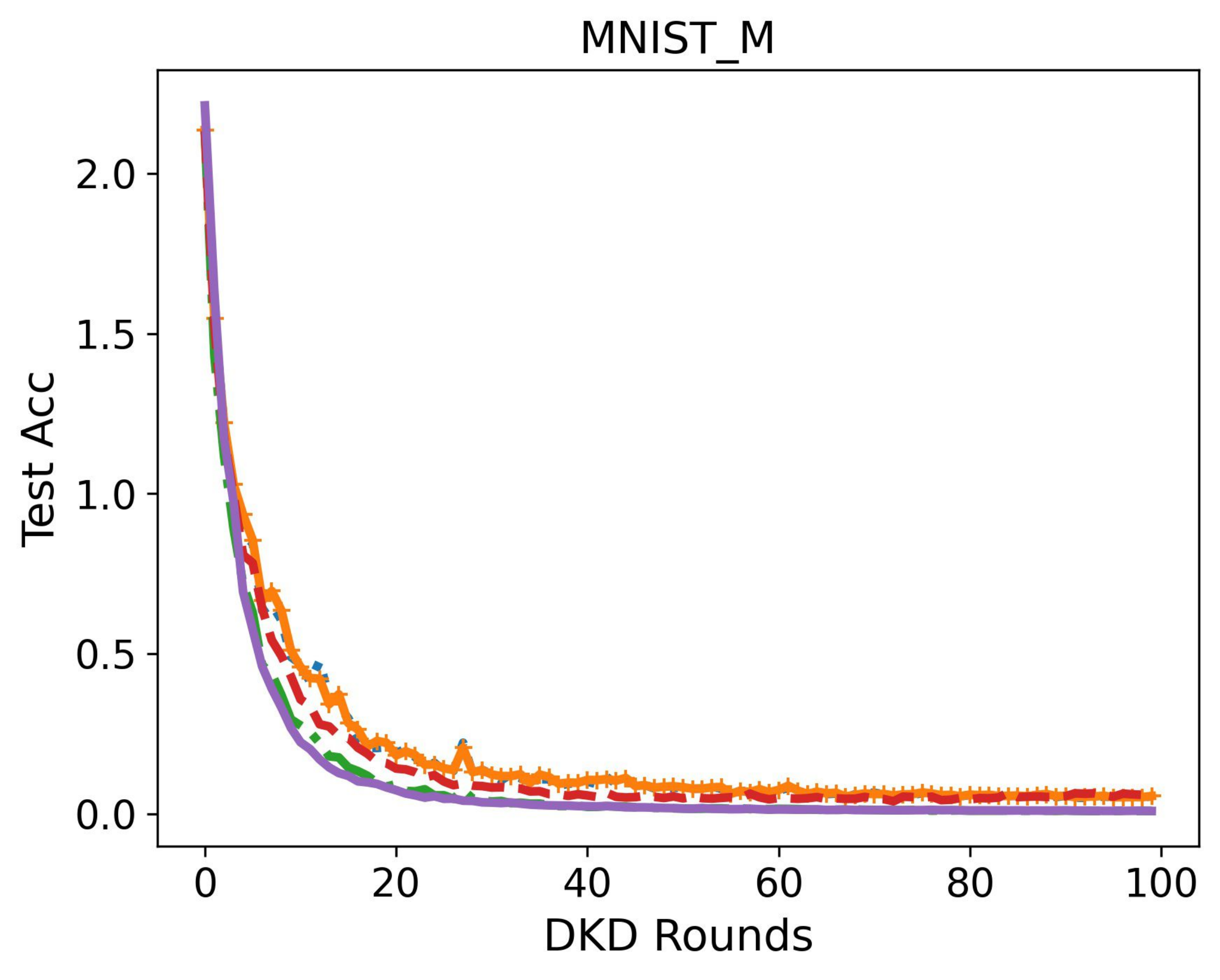}
    \includegraphics[width=0.45\columnwidth]{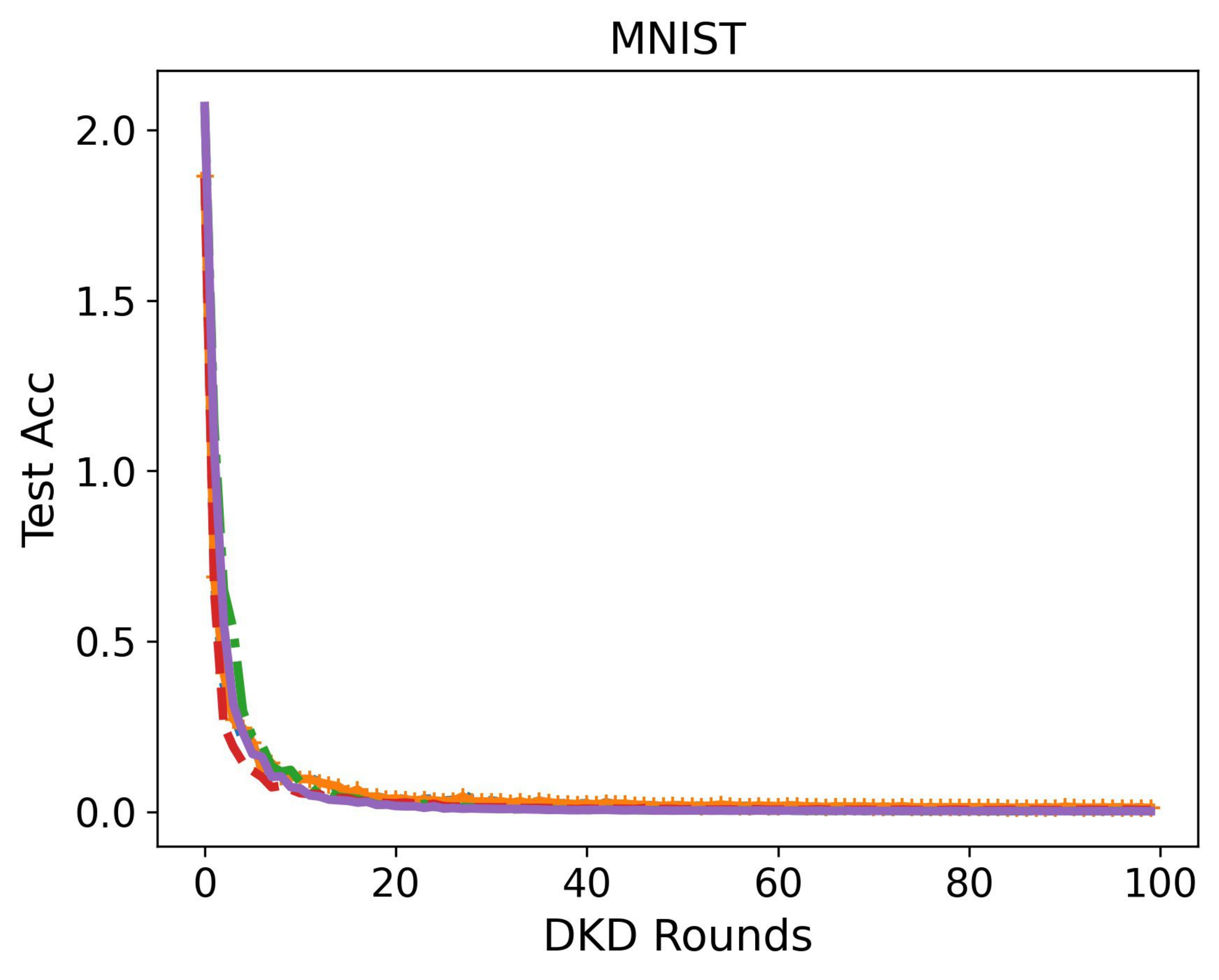}
    \caption{Convergence of the training loss for the multi-sources digits datasets.}
    \label{fig::fedbn_train_loss}
\end{figure}

\textbf{Additional Results}
Figure \ref{fig::fedbn_train_loss} illustrates the training loss and the number of DKD rounds for each dataset.
\end{document}